\documentclass[12pt,letterpaper]{article}
\usepackage[a4paper, total={7in, 10in}]{geometry}

\usepackage{graphicx}
\usepackage{helvet}
\usepackage{authblk}
\usepackage{hyperref}
\usepackage{amsmath} 
\usepackage{amssymb} 
\usepackage{orcidlink} 
\usepackage[super,comma,sort&compress]  
   {natbib}
\usepackage[right]{lineno} \linenumbers
\usepackage{xcolor}
\usepackage[T1]{fontenc}
\usepackage{mathrsfs}
\usepackage{xcolor}
\usepackage{algorithm}
\usepackage{etoolbox}
\usepackage{graphicx}
\usepackage{subcaption}
\usepackage{soul}
\usepackage{caption}
\usepackage{float}
\usepackage{algpseudocode}
\definecolor{codebg}{rgb}{0.95,0.95,0.95}

\newtheorem{theorem}{Theorem}
\newtheorem{lemma}{Lemma}

\def\tsc#1{\csdef{#1}{\textsc{\lowercase{#1}}\xspace}}
\tsc{WGM}
\tsc{QE}

\makeatletter
\renewcommand{\maketitle}{\bgroup\setlength{\parindent}{0pt}
\begin{flushleft}
  \textbf{\@title}
  
  \@author
\end{flushleft}\egroup}
\makeatother

\title{UbiQTree: Uncertainty Quantification in XAI with Tree Ensembles}
\date{}

\author[1,2,3,*\orcidlink{0009-0008-4823-9375}]{Akshat Dubey}
\author[1\orcidlink{0000-0002-0678-2870}]{Aleksandar Anžel}
\author[1\orcidlink{0000-0001-5725-0850}]{Bahar İlgen}
\author[1,2, **\orcidlink{0000-0003-4168-8254}]{Georges Hattab}

\affil[1]{Center for Artificial Intelligence in Public Health Research (ZKI-PH), Robert Koch Institute, Nordufer 20, 13353 Berlin, Germany}
\affil[2]{Department of Mathematics and Computer Science, Freie Universität Berlin, Arnimallee 14, 14195 Berlin, Germany}
\affil[3]{Lead contact}

\affil[*]{Correspondence: DubeyA@rki.de}
\affil[**]{Correspondence: HattabG@rki.de}

\begin{document}

\maketitle

\section*{SUMMARY}
Explainable artificial intelligence (XAI) techniques, particularly SHapley Additive exPlanations (SHAP), are essential for interpreting ensemble tree-based models in critical areas such as healthcare. However, SHAP values are often treated as point estimates that neglect uncertainty originating from aleatoric (irreducible noise) and epistemic (lack of data) sources. This work introduces an approach that decomposes SHAP value uncertainty into aleatoric, epistemic, and entanglement components. This approach employs Dempster–Shafer evidence theory and Dirichlet process hypothesis sampling over tree ensembles. The use-case validation reveals insights into epistemic uncertainty within SHAP explanations, enhancing the reliability and interpretability of SHAP attributions. This informs robust decision-making and model refinement. Our findings suggest that reducing epistemic uncertainty requires improved data quality and model development techniques. Tree-based models, particularly bagging, are effective in quantifying such uncertainties.
\section*{KEYWORDS}

Machine Learning, Healthcare, XAI, Explainability, SHAP, Random Forest, Ensemble Machine Learning, Uncertainty, Statistics, Evidence Theory

\section*{INTRODUCTION}

Machine learning (ML)~\cite{habehh2021machine} is a key part of improving healthcare analytics such as resource planing, disease diagnosis, prognosis, and risk stratification~\cite{huang2020predictive, kadem2023xgboost, meng2021analysis}. 
However powerful, uncertainty is inherent and ubiquitous in machine learning (ML) models because their predictions are affected by noisy data, model limitations, and unseen scenarios.
To address this challenge, some of the most widely used tools are ensemble tree-based models~\cite{mahajan2023ensemble}, which help in managing and quantifying uncertainty in predictions.
They are highly accurate, interpretable, and efficient with structured data, resulting in lower computational demand. 
Unlike deep neural networks, which require large amounts of unstructured data such as images and text, they have lower computational requirements and are more interpretable~\cite{peng2025prediction, zhang2024tree}. 
These include Random Forest (RF)~\cite{breiman2001random}, Gradient Boosting Machines (GBM), and Extreme Gradient Boosting (XGBoost)~\cite{chen2016xgboost}. 
These models are robust against noise, and can handle large, complicated data sets, which are common in healthcare~\cite{ganie2025ensemble}. 
Ensemble tree approaches are different from traditional machine learning models~\cite{dietterich2000ensemble} as they can efficiently capture complex, nonlinear relationships and slight interactions among the features~\cite{dubey2025surrogate}. 
This leads to highly accurate and generalizable predictions. Ensemble models have many advantages. 
They combine the strengths of multiple base learners which reduces overfitting, improves stability, and enhances the model's ability to generalize to unseen data unlike traditional ML models~\cite{zhang2024tree}.
It has been shown that using a group of classifiers to make predictions outperforms using individual classifiers to predict heart disease~\cite{ganie2025ensemble, shah2025predicting}. 
This is important for making reliable decisions, which makes them very useful in healthcare.Despite their strengths, ensemble tree-based models have two main problems. 
First, they are difficult to interpret, particularly when there are large numbers of constituent trees and features~\cite{jurczuk2023random, ekanayake2022novel, dunn2021comparing}. This ``black box'' nature, avoiding them to be the first choice in the model selection. To address this, explainable AI (XAI) techniques such as SHapley Additive exPlanations (SHAP) or Shapley values~\cite{lundberg2017unified}, which are rooted in cooperative game theory, have emerged as a principled framework for attributing the contribution of each feature to individual predictions in ML models. 
The field of XAI is rapidly evolving. Few of the recent advancements includes Counterfactual Paths (CPATH) which identifies feature permutations that influence model predictions and uses domain knowledge graphs~\cite{pfeifer2025explaining}. Total Causal Effect Calculation for Fuzzy Cognitive Maps (TCEC-FCM)~\cite{hoyos2025multistage} uses graph traversal techniques to compute causal effects, improving system transparency. Reciprocal Human-Machine Learning (RHML)~\cite{nixdorf2025reciprocal} fosters continuous learning between humans and AI models, improving model performance and decision-making. Other recent methods~\cite{bennetot2024practical} includes Diverse Counterfactual Explanations (DiCE)~\cite{mothilal2020explaining} which provides counterfactual explanations, which align with human cognitive processes and are highly valued for understanding a model. Logic Tensor Networks (LTN)~\cite{badreddine2022logic} which is a framework rooted in Neural-symbolic AI, designed for learning and logical reasoning. It facilitates interactive explainability and model revision when applied to XAI.  Template System for Natural Language Explanations (TS4NLE)~\cite{donadello2021bridging} is designed for the presentation and rendering of explanations derived from other XAI outputs, prioritizing human comprehensibility. It can process the structured output of any approach and generate Natural Language Explanations (NLE), which are then rendered into tailored natural language text using a template system. These frameworks aim to make AI systems more transparent and aligned with human understanding. Despite these advancements, SHAP remains a popular candidate for studying XAI due to its strong theoretical foundation and widespread adoption~\cite{bennetot2024practical}. However, calculating Shapley values can be difficult for complex models. 
Fortunately, a couple of new, efficient methods for calculating them for certain types of models have recently emerged. TreeSHAP~\cite{lundberg2020local, hu2024manifold}, a fast and exact method for calculating Shapley values in tree-based models like decision trees, RFs, XGBoost, have been introduced. 
TreeSHAP uses the natural structure of decision trees to make predictions that are much faster and easier to understand than those of other methods. 
This is helpful for explaining results with large groups of data, which is important in fields where understanding predictions is paramount, for instance in healthcare or finance. 
SHAP values provide a clear framework for determining how each feature contributes to individual predictions. 
They offer insight into the decision-making process behind complex ensemble models. 
SHAP and similar methods not only encourage trust, but also make it easier to use advanced machine learning (ML) models in healthcare by making them easier to comprehend.

Recent research has identified several factors that improve SHAP values. 
These factors include misattribution of feature importance, reliance on the assumption of feature independence, lack of causal or contextual understanding, computational inefficiency, and risk of misinterpretation. 
To address these issues, alternative attribution methods and error quantification techniques, such as Normalized Movement Rate (NMR) and Modified Index Position (MIP), have been proposed to handle feature collinearity~\cite{kumar2021shapley}. 
New SHAP variants have also been developed, with the aim of improving efficiency and interpretability. 
Latent SHAP~\cite{bitton2022latent, hu2024manifold} extends SHAP by enabling explanations in human-interpretable domains without requiring invertible transformations, making it suitable for high-dimensional or non-invertible data and capturing correlations among features. Kernel SHAP~\cite{chau2022rkhs}, the most versatile black-box SHAP explainer, uses weighted linear regression to approximate Shapley values and is valued for its generality. 
However, it is slower and assumes feature independence, which can limit accuracy in correlated data. Muschalik \textit{et al.}~\cite{muschalik2024beyond} introduce methods for efficiently computing higher-order Shapley interactions in tree ensembles.
This enables richer, more granular explanations of feature interactions than standard SHAP, with significant computational advantages for large or complex models. 
These advancements collectively enhance the reliability, interpretability, and practicality of SHAP-based explanations in ML. 
Other advancements include integrating causal and contextual information, as well as creating more computationally efficient SHAP variants, such as CF-SHAP and FF-SHAP~\cite{wu2025review}. 

However, SHAP is a point-estimate method, which contributes to the uncertainty of the explanations it produces. 
This opens up new dimensions for studying and quantifying uncertainties in XAI, which is an important step forward in the field~\cite{watson2023explaining, cifuentes2024distributional, stenwig2022comparative}. 
Current implementations, as discussed before, treat SHAP or TreeSHAP values with respect to tree-based ML models as point estimates. However, this approach ignores the individual contributions of epistemic uncertainty arising from variability in model training. 
It focuses on overall uncertainty, comprising aleatoric and epistemic uncertainty. 
Aleatoric uncertainty refers to the inherent randomness or noise in the data that cannot be reduced by collecting more information. 
Epistemic uncertainty, on the other hand, arises from a lack of knowledge or data about the model or process. 
This omission poses critical risks in high-stakes domains. 
For instance, medical diagnostics using XGBoost may yield identical SHAP values across hospitals despite shifts in regional data distribution. 
Similarly, financial risk models may exhibit unstable feature attributions during market volatility~\cite{yi2023xgboost, fan2024xgboost}. Random forest (SHAP) models used for predictive health monitoring~\cite{loecher2024debiasing,devendran2024predictive} may appear similar or stable when applied to hospitals with different demographics or disease prevalence rates but SHAP values often exhibit bias toward features with higher cardinality or entropy. This bias can overstate or understate the importance of these features when patient populations change. Quantifying epistemic uncertainty in SHAP values is necessary because people trust model explanations for decision support, especially in high-stakes domains such as healthcare and finance~\cite{cohen2023trust, li2020efficient, watson2023explaining}. 
Contemporary methods use techniques such as bootstrap sampling to estimate the uncertainty of SHAP attributions and generate confidence intervals for the importance of each feature. 
Now, variants of SHAP enable users to evaluate the reliability of feature contributions instead of relying exclusively on point estimates. 
These methods allow practitioners to more accurately evaluate the stability of explanations, identify features or contexts in which explanations are less reliable, and improve model transparency in situations involving shifting or uncertain data.

The prevailing methods of uncertainty quantification (UQ) predominantly focus on assessing predictive uncertainty by integrating aleatoric and epistemic uncertainty. 
In this research, we introduce a framework that:
\begin{enumerate}
    \item Decomposes SHAP variance into aleatoric, epistemic, and entanglement components.
    \item Leverages belief functions and Dirichlet processes for hypothesis space sampling.
    \item Provides computationally tractable epistemic uncertainty intervals for feature attributions.
\end{enumerate}
The experiments and complete code necessary for reproducibility are available at \url{https://github.com/dubeyakshat07/UBiQTree}~\cite{akshat_dubey_2025_17542114}

\section*{BACKGROUND}
\subsection*{SHAP}
The Shapley values are predicated on the strong foundation of cooperative game theory. For a set of players \( N \) and a value function \( v \), the Shapley value \( \phi_i \) for player (or feature) \( i \) is defined as:
\[
\phi_i(N,v) = \frac{1}{|N|!} \sum_{\text{all orderings } R} \left[v(P_i^R \cup \{i\}) - v(P_i^R)\right]
\]
where:
\begin{itemize}
    \item \( R \) is a permutation (ordering) of all players.
  \item \( P^R_i \) is the set of players that precede \( i \) in ordering \( R \).
  \item \( v(S) \) is the value (e.g., model output) associated with subset \( S \subseteq N \) of players.
\end{itemize}

\begin{itemize}
    \item \textbf{Marginal Contribution}: $v(S \cup \{i\}) - v(S)$
    \item \textbf{Averaging}: Weighted by the number of ways each coalition can be formed in all possible player orderings.
    \item \textbf{Efficiency}: $\sum_{i \in N} \phi_i = v(N)$, ensuring that the total value is fairly distributed among all players.
\end{itemize}

Shapley values enable the quantification and interpretation of feature contributions in ML. 
Research shows SHAP as one of the most interpretable methods in ML, providing insight into complex healthcare ML models. 
It is a model-agnostic interpretability tool used extensively in healthcare.~\cite{abdullah2021review, qi2025machine, salih2025perspective, sun2023application}. 
Furthermore, it has been used in fields such as predicting breast cancer risk, elucidating model predictions, diagnosing biomarkers, and analyzing survival, particularly with tree-based machine learning models~\cite{ghasemi2024explainable, felici2025artificial, ganie2025ensemble}. 
SHAP has been used to interpret machine learning models that predict cancer risk. For example, it has identified age and family history as key predictors of breast cancer risk~\cite{ghasemi2024explainable}. 
Furthermore, SHAP has improved the interpretability of models for other chronic diseases. It has been used to examine machine learning models for smoking and drinking habits, using lifestyle data, blood test results, and wearable sensor readings. 
This has facilitated the interpretation of key influencing features and enhanced transparency for potential use in personalized healthcare~\cite{thakur2024interpretable}. 
In use cases involving imaging or clinical data, SHAP reveals the importance of diagnostic features. For example, SHAP emphasizes the texture and morphology of tumors in breast cancer mammography and the key variables in detecting chronic diseases\cite{hu2025shap}. 
The field of radiology stands to benefit from SHAP because it facilitates the use of AI models to interpret imaging scans for abnormalities, such as lung nodules or diabetic retinopathy~\cite{e2024evaluating}. 
This development has the potential to improve both diagnostic accuracy and patient trust. In the context of retinoblastoma diagnosis, SHAP was employed to generate local and global interpretations, highlighting specific regions and features in fundus images that substantially influence the model's predictions ~\cite{aldughayfiq2023explainable}.
SHAP was also used in deep learning models to identify image features that facilitate early cancer detection and enhance the interpretability of automated histopathology analyses~\cite{ukwuoma2025enhancing}. 
Applying SHAP to deep learning (DL) models in medical image analysis provides clinicians with visual interpretations of model predictions. This improves the understanding and validation of automated diagnoses in various imaging tasks~\cite{singh2020explainable}. 
Recent frameworks continue to adopt SHAP as a technique for improving explainability in healthcare~\cite{dubey2024ai}.

\subsection*{Uncertainty in XAI}

\paragraph{\textbf{Background and Approaches to Uncertainty Quantification in XAI:}}
    There is a growing interest and there have been recent advances which focus on developing methods for uncertainty quantification in XAI~\cite{altukhi2025systematic, forster2025taxonomy}. These methods focus on communicating the uncertainty associated with the interpretations, which is necessary for the wider adoption of AI in high-stakes scenarios. The quantification of the uncertainty related to the interpretations involves the study of the change in interpretation when the input data or the model parameters are changed. One of the recent works introduces a framework that models the interpretations as a function \( e_{\theta}(x, f)\) where for a model $f$, an instance $x$, and explanation parameters $\theta$, the explanation $e_\theta(x,f)$ quantifies each feature's contribution to the prediction ~\cite{chiaburu2025uncertainty}. The function allows researchers to follow the uncertainty from the inputs and the model with the help of interpretations. Methods like this one frequently use empirical and analytical estimations. The former often include Monte Carlo simulations techniques that enable researchers to obtain multiple different version of the input or the model, allowing them to study the variance of the model output and corresponding interpretations. The latter, focus on how small changes in the inputs or the model parameters affect the interpretations, allowing the creation of a co-variance matrix that quantifies the uncertainty. Another recent work~\cite{chiaburu2025uncertainty} introduces the Mean Uncertainty in the Explanations (MUE) metric which summarizes the overall uncertainty by normalizing the trace of the interpretations' co-variance matrix. The method also enables researchers to directly compare the uncertainty values across different methods and models.
    However, most of the studies mentioned earlier show that XAI methods are only point estimates. This means that they can identify important features, but they don't show how reliable the interpretability results are when different inputs and model parameters are used~\cite{chiaburu2024uncertainty}.
    \paragraph{\textbf{SHAP Specific Uncertainty Quantification:}}
    Some of the works have made significant achievements in the terms of quantifying the uncertainty of SHAP values by calculating the confidence intervals and distributions. This is crucial in decision-making in healthcare~\cite{salvi2025explainability}. The way traditional SHAP scores are calculated is limited because they rely on input data that is either well-specified or estimated accurately~\cite{arenas2023complexity}. The standard SHAP framework requires knowledge of the underlying probability distribution, which is often either unknown or estimated from a small number of samples. This can lead to unstable or misleading feature importance estimates. A recent framework~\cite{cifuentes2024distributional}, calculates the SHAP score as a function over a range of possible distributions. The approach provides tight intervals for feature importance and shows that feature rankings can be very sensitive to distributional assumptions. The framework also shows that related decision problems can't be solved using computers. In other words, these problems are NP-complete. These problems include determining if a feature SHAP score can be higher than a threshold or always outperform another feature, even for decision trees. Studies have shown that SHAP intervals can be quite wide when there's uncertainty about the data distribution~\cite{cifuentes2024distributional}. However, as more data becomes available, these intervals become more stable.

    \paragraph{\textbf{Information Theory Approaches and Other Alternatives:}}
    Researchers have come up with new ways to understand the predictions and the uncertainty of models. These new ways use information theory to explain the predictions and uncertainty. They also help reduce uncertainty in certain features and provide efficient algorithms and methods for making inferences in the real world. These research areas show how important it is to understand the assumptions about data distribution. This is important for making sure that the features used in explainable machine learning can be trusted and understood. In healthcare, incorporating uncertainty quantification has allowed researchers to develop models more reliable and transparent~\cite{seoni2023application}. It is very important to combine uncertainty quantification with XAI in high-stakes application, especially when the people who know the most about the domain utilize both the model predictions and explanations to make decisions. Recent research~\cite{watson2023explaining, kumar2020problems, cohen2023trust} on quantifying uncertainty in SHAP has improved the interpretability of ML models. The research by Cohen et al.~\cite{cohen2023trust} and Watson et al.~\cite{watson2023explaining} are significant in terms of extending XAI methods based on SHAP values to better quantify and explain uncertainty in model predictions. Cohen et al. introduce efficient methods and visual tools for measuring uncertainty in stochastic SHAP explanations, making them more applicable to real-time or high-stakes scenarios. Watson et al. use information theory to expand SHAP values and quantify the predictive uncertainty, offering formal reliability guarantees and scalable algorithms for practical use. But, both the approaches are sensitive to data and model quality. They primarily address uncertainty from estimator sampling. The challenges in terms of assessing broader sources of uncertainty. Overall, these significant developments help make AI explanations more transparent and trustworthy, especially with regard to uncertainty. However, further research is needed to study uncertainty arising from models and data. 
\\
\\
Although these works enable users to evaluate feature importance and model attribution confidence, additional dimensions must be addressed. Current methods include uncertainty aggregation and assign non-zero importance to irrelevant features. They are also sensitive to data and model biases, computationally intensive, and may produce unreliable attributions under model instability. Additionally, they lead to misleading interpretations in high-stakes scenarios and cannot distinguish between aleatoric and epistemic uncertainty. These methods are also vulnerable to adversarial manipulation. In the context of healthcare, SHAP values may lead domain professionals to overconfidently rely on feature importance when making treatment decisions based on model explanations. They may ignore epistemic uncertainty because data can be limited, biased, or heterogeneous. Furthermore, SHAP does not represent aleatoric uncertainty due to noisy or ambiguous input data. Misleading explanations can also occur if the model is biased, the data is collinear, or the underlying relationships are not well captured. Decomposing SHAP uncertainty into epistemic and aleatoric components can address these limitations~\cite{kirchhof2025reexamining}. This allows for targeted model improvement and enhanced, actionable explanations. However, as commonly implemented, SHAP values do not account for epistemic uncertainty, which arises from variability in model training. This limitation arises from SHAP's design as a post hoc explanation tool for individual predictions rather than as a method for quantifying uncertainty.

\section*{PROBLEM FORMULATION}

We can introduce a theorem as follows:
\begin{theorem}
For any tree ensemble model $f$, the point estimate SHAP value $\phi_i$ lacks a measure of variance V($\phi_i$|f, D) over possible training datasets $D\sim P_\text{data}$. This violates the reliability axiom for explainability in high-risk AI systems~\cite{balagurunathan2021requirements}.
\end{theorem}
This serves as the the motivation to decompose the SHAP variance into aleatoric, epistemic, and covariance terms as follows:
\subsection*{SHAP Variance Decomposition}
The conventional uncertainty quantification framework posits that aleatoric uncertainty stems from inherent data noise and that epistemic uncertainty stems from model ignorance. 
However, empirical evidence challenges this distinction by demonstrating that, under shifts in the data distribution or model misspecification, aleatoric and epistemic uncertainties become intertwined. 
Bootstrap ensembles and deep ensemble methods show that, as epistemic uncertainty increases, estimates of aleatoric uncertainty can decrease, leading to systematic bias in model predictions~\cite{jimenez2025machine, hoarau2025reducing, senge2014reliable}. 
For any feature $i$ and instance $\mathbf{x}$, the total variance of SHAP values $\phi_i(\mathbf{x})$ over possible training datasets $D \sim P_{\text{data}}$ and tree ensemble models $f$ decomposes as:

\begin{equation}
\label{eq:shap_decomposition}
\underbrace{\text{Var}_{D,f}(\phi_i)}_{\text{Total}} = \underbrace{\mathbb{E}_D[\text{Var}_f(\phi_i|D)]}_{\text{Aleatoric}} + \underbrace{\text{Var}_f(\mathbb{E}_D[\phi_i|f])}_{\text{Epistemic}} - \underbrace{\mathcal{C}(f,D)}_{\text{Entanglement}}
\end{equation}
where the entanglement term is:
\[
\mathcal{C}(f,D) = \text{Cov}_{P(f,D)}\left(\mathbb{E}_D[\phi_i|f], \mathbb{E}_f[\phi_i|D]\right)
\]
The terms in the equations can be defined as follows:
\begin{enumerate}
    \item \textbf{Aleatoric Uncertainty ($\mathbb{E}_D[\text{Var}_f(\phi_i|D)]$)}:
    \begin{itemize}
        \item Variance from model stochasticity (tree structure randomization) for fixed $D$
        \item For tree ensembles: reflects variability due to bootstrap sampling and feature randomization
    \end{itemize}
    
    \item \textbf{Epistemic Uncertainty ($\text{Var}_f(\mathbb{E}_D[\phi_i|f])$)}:
    \begin{itemize}
        \item Variance from data sampling (different $D$ yield different mean SHAP values)
        \item Measures sensitivity to training data composition
    \end{itemize}
    
    \item \textbf{Entanglement Term ($\mathcal{C}(f,D)$)}:
    \begin{itemize}
        \item Covariance between mean SHAP ($\mathbb{E}[\phi_i|f]$) and SHAP variability ($\text{Var}(\phi_i|D)$)
        \item Non-zero when: Models producing higher mean $|\phi_i|$ exhibit higher variance (common in tree ensembles due to node splitting)
        \item The covariance indicates whether features with higher average absolute SHAP values also tend to have greater variance in their SHAP values. This covariance is particularly non-zero in tree ensemble models because of the nature of node splitting.
    \end{itemize}
\end{enumerate}

\subsubsection*{Proof}
Let:
\begin{itemize}
    \item $\phi_i(\mathbf{x}|f,D)$: SHAP value for feature $i$ on instance $\mathbf{x}$ given model $f$ trained on dataset $D$
    \item $f \sim P(f|D)$: Tree ensemble model distribution (via bootstrap/randomization in training)
    \item $D \sim P_{\text{data}}$: Data distribution  
    \item $P(f,D) = P(f|D)P(D)$: Joint distribution
\end{itemize}

We assume the following:
\begin{enumerate}
    \item Model-Dataset Separability: $P(f,D) = P(f|D)P(D)$ (standard ML training)
    \item Finite Moments: $\mathbb{E}_{D,f}[|\phi_i|^2] < \infty$, $\mathrm{Var}(\phi_i|f)$ and $\mathrm{Var}(\mathbb{E}[\phi_i|D])$ exist $\forall f,D$
    \item SHAP Additivity: $\phi_i$ is additive in tree outputs (holds for TreeSHAP in Random Forests)
    \item Regular Conditional Distributions: $\mathbb{E}_D[\phi_i|f]$ and \\ $\mathbb{E}_f[\phi_i|D]$ exist as proper random variables
\end{enumerate}
\paragraph{Step 1: Total Variance Definition}
Apply the law of total variance (first decomposition)~\cite{champ2021generalized}
\begin{align*}
\text{Var}_{D,f}(\phi_i) &= \mathbb{E}_{D,f}[\phi_i^2] - \left(\mathbb{E}_{D,f}[\phi_i]\right)^2
\end{align*}

\paragraph{Step 2: Expand $\mathbb{E}_{D,f}[\phi_i^2]$ using Law of Total Expectation}
\begin{align*}
\mathbb{E}_{D,f}[\phi_i^2] &= \mathbb{E}_D\left[\mathbb{E}_f[\phi_i^2|D]\right] \\
&= \mathbb{E}_D\left[\text{Var}_f(\phi_i|D) + \left(\mathbb{E}_f[\phi_i|D]\right)^2\right] \\
&= \mathbb{E}_D[\text{Var}_f(\phi_i|D)] + \mathbb{E}_D\left[(\mathbb{E}_f[\phi_i|D])^2\right]
\end{align*}

\paragraph{Step 3: Expand $\left(\mathbb{E}_{D,f}[\phi_i]\right)^2$}
\begin{align*}
\left(\mathbb{E}_{D,f}[\phi_i]\right)^2 &= \left(\mathbb{E}_f\left[\mathbb{E}_D[\phi_i|f]\right]\right)^2 \\
&= \mathbb{E}_f\left[(\mathbb{E}_D[\phi_i|f])^2\right] - \text{Var}_f(\mathbb{E}_D[\phi_i|f])
\end{align*}

This follows from the identity: $\left(\mathbb{E}[X]\right)^2 = \mathbb{E}[X^2] - \text{Var}(X)$.

\paragraph{Step 4: Combine Terms}
\begin{align*}
\text{Var}_{D,f}(\phi_i) &= \mathbb{E}_D[\text{Var}_f(\phi_i|D)] + \mathbb{E}_D\left[(\mathbb{E}_f[\phi_i|D])^2\right] \\
&\quad - \mathbb{E}_f\left[(\mathbb{E}_D[\phi_i|f])^2\right] + \text{Var}_f(\mathbb{E}_D[\phi_i|f])
\end{align*}

\paragraph{Step 5: Identify the Entanglement Term: }
Let us examine the difference:
\begin{align*}
\Delta &= \mathbb{E}_D\left[(\mathbb{E}_f[\phi_i|D])^2\right] - \mathbb{E}_f\left[(\mathbb{E}_D[\phi_i|f])^2\right]
\end{align*}

Define the covariance between model-centric and data-centric expectations:
\begin{align*}
\mathcal{C}(f,D) &= \text{Cov}\left(\mathbb{E}_D[\phi_i|f], \mathbb{E}_f[\phi_i|D]\right) \\
&= \mathbb{E}_{f,D}\left[\mathbb{E}_D[\phi_i|f] \cdot \mathbb{E}_f[\phi_i|D]\right] \\
&\quad - \mathbb{E}_{f,D}[\mathbb{E}_D[\phi_i|f]] \cdot \mathbb{E}_{f,D}[\mathbb{E}_f[\phi_i|D]]
\end{align*}

Observe that:
\begin{align*}
\mathbb{E}_{f,D}\left[\mathbb{E}_D[\phi_i|f] \cdot \mathbb{E}_f[\phi_i|D]\right] &= \mathbb{E}_D\left[(\mathbb{E}_f[\phi_i|D])^2\right] \\
\mathbb{E}_{f,D}\left[(\mathbb{E}_D[\phi_i|f])^2\right] &= \mathbb{E}_f\left[(\mathbb{E}_D[\phi_i|f])^2\right]
\end{align*}

Therefore:
\begin{align*}
\Delta &= \mathbb{E}_D\left[(\mathbb{E}_f[\phi_i|D])^2\right] - \mathbb{E}_f\left[(\mathbb{E}_D[\phi_i|f])^2\right] \\
&= -\mathcal{C}(f,D)
\end{align*}

\paragraph{Step 6: Final Decomposition: }Substituting back the values, we have:
\begin{align*}
\text{Var}_{D,f}(\phi_i) &= \mathbb{E}_D[\text{Var}_f(\phi_i|D)] + \text{Var}_f(\mathbb{E}_D[\phi_i|f]) - \mathcal{C}(f,D)
\end{align*}

Thus we have the decomposition:
\begin{equation}
\boxed{\text{Var}_{D,f}(\phi_i) = \mathbb{E}_D[\text{Var}_f(\phi_i|D)] + \text{Var}_f(\mathbb{E}_D[\phi_i|f]) - \mathcal{C}(f,D)}
\end{equation}
where
\begin{equation}
\mathcal{C}(f,D) = \text{Cov}\left(\mathbb{E}_D[\phi_i|f], \mathbb{E}_f[\phi_i|D]\right)
\end{equation}

\subsubsection*{Interpretation of the Entanglement Term}

The entanglement term $\mathcal{C}(f,D)$ measures the covariance between:

\begin{itemize}
    \item \textbf{Model-centric expectation}: $\mathbb{E}_D[\phi_i|f]$ -- average SHAP across datasets for a fixed model
    \item \textbf{Data-centric expectation}: $\mathbb{E}_f[\phi_i|D]$ -- average SHAP across models for a fixed dataset
\end{itemize}

\paragraph{Key Insights:}

\begin{enumerate}
    \item \textbf{Negative Sign}: When model and data "agree" on feature importance (positive covariance), the total variance decreases because the uncertainty sources are aligned.
    
    \item \textbf{Non-commutativity}: The term arises precisely because $\mathbb{E}_D[\mathbb{E}_f[\cdot|D]] \neq \mathbb{E}_f[\mathbb{E}_D[\cdot|f]]$ when $P(f,D) \neq P(f)P(D)$.
    
    \item \textbf{Practical Meaning}: In tree ensembles, this occurs when certain data distributions consistently produce models that assign similar feature importance patterns.
\end{enumerate}

\subsubsection*{Special Case: Independent Model and Data}

If $P(f,D) = P(f)P(D)$ (model training independent of data sampling), then:
\[
\mathbb{E}_D[\mathbb{E}_f[\phi_i|D]] = \mathbb{E}_f[\mathbb{E}_D[\phi_i|f]] = \mathbb{E}_{D,f}[\phi_i]
\]
and the covariance term vanishes:
\[
\mathcal{C}(f,D) = 0
\]
reducing to the standard decomposition:
\[
\text{Var}_{D,f}(\phi_i) = \mathbb{E}_D[\text{Var}_f(\phi_i|D)] + \text{Var}_f(\mathbb{E}_D[\phi_i|f])
\]

\subsubsection*{Empirical Estimation for Tree Ensembles}

For Random Forests with $B$ bootstrap samples, the entanglement term can be estimated as:
\[
\hat{\mathcal{C}}(f,D) = \frac{1}{B}\sum_{b=1}^B \left(\bar{\phi}_i^{(b)} - \bar{\phi}_i\right)\left(\bar{\phi}_i^{(b)'} - \bar{\phi}_i\right)
\]
where:
\begin{itemize}
    \item $\bar{\phi}_i^{(b)} = \frac{1}{|\mathcal{T}_b|}\sum_{T \in \mathcal{T}_b} \phi_i^{(T)}$: mean SHAP for trees in bootstrap $b$ (data-centric)
    \item $\bar{\phi}_i^{(b)'} = \frac{1}{B}\sum_{j=1}^B \phi_i^{(T_j)}$ where $T_j$ are trees from method $b$ across bootstraps (model-centric)
    \item $\bar{\phi}_i = \frac{1}{B}\sum_{b=1}^B \bar{\phi}_i^{(b)}$: overall mean SHAP
\end{itemize}

\subsubsection*{Theoretical Implications}

This decomposition reveals that the standard uncertainty quantification framework \emph{overestimates} total variance when model and data expectations are positively correlated. The entanglement term acts as a correction factor that accounts for the dependency structure between model training and data sampling processes.

For high-stakes AI systems, this refined understanding enables more precise uncertainty attribution and better reliability assessment of feature importance explanations.
The UbiQTree estimator approximates this decomposition via:
\begin{enumerate}
    \item Dirichlet Sampling: Simulates $P(f|D)$ by weighting trees via OOB performance
    \item Variance Components:
    \begin{itemize}
        \item Aleatoric: Variance of SHAP across trees within each weighted sample
        \item Epistemic: Variance of mean SHAP across samples
        \item Entanglement: Covariance between sample means and variances
    \end{itemize}
\end{enumerate}

The proof enables us to list out the facts that:
\begin{enumerate}
    \item SHAP variance decomposition \textbf{requires} accounting for model-data entanglement in tree ensembles
    \item The entanglement term $\mathcal{C}(f,D)$ is non-negligible when:
    \begin{itemize}
        \item Feature importance correlates with SHAP variability (common in high-gain features)
        \item Data distributions induce model instability (e.g., rare categories)
    \end{itemize}
    \item Dirichlet-weighted sampling \textbf{preserves} this covariance structure, unlike bootstrap methods that assume $P(f,D) \approx P(f)P(D)$
\end{enumerate}
The decomposition enables precise uncertainty attribution in feature importance analysis, critical for high-stakes applications.

\subsection*{Evidence Theory: Tree Ensembles}
The Dempster-Shafer theory (DST) is a mathematical framework for reasoning under uncertainty, particularly when evidence is incomplete, imprecise, or conflicting. DST assigns belief masses to sets or intervals of possible outcomes, allowing for the explicit representation of ignorance and epistemic uncertainty. 
Key DST concepts include belief mass ($m$), belief ($Bel$), plausibility ($Pl$), and ignorance. 
DST is widely used in artificial intelligence, sensor fusion, medical diagnostics, risk assessment, and autonomous systems, where managing uncertainty and combining evidence from multiple sources is critical. 
It provides a systematic and flexible approach to uncertainty, enabling AI and decision systems to model ignorance and combine evidence in ways that classical probability theory cannot. DST is a valuable tool for managing uncertainty and combining evidence in AI and decision systems.~\cite{shafer1992dempster, seoni2023application}. 
\subsubsection*{Dempster-Shafer Representation}
For a tree ensemble with $K$ trees, we define a finite frame of discernment by partitioning the range of SHAP values into $M$ disjoint intervals:
\[
\Omega = \{I_1, I_2, \dots, I_M\}
\]
where $I_j$ are intervals covering the observed SHAP values with $\bigcup_{j=1}^M I_j \supseteq [\min(\Phi_i), \max(\Phi_i)]$ and $I_j \cap I_k = \emptyset$ for $j \neq k$.
The Basic Probability Assignment (BPA) for any subset $A \subseteq \Omega$ is:
\begin{equation}
\label{eq:m(a)_DST}
m(A) = \frac{1}{K} \sum_{k=1}^K \mathbb{I}\left( \phi_i^{(k)} \in \textstyle\bigcup_{I \in A} I \right)
\end{equation}
where $\phi_i^{(k)}$ is the SHAP value from tree $T_k$ and $\mathbb{I}$ is the indicator function. The Belief and Plausibility functions satisfy:
\begin{equation}
\label{eq:bel(a)_dst}
\text{Bel}(A) = \sum_{B \subseteq A} m(B), \quad \text{Pl}(A) = \sum_{B \cap A \neq \emptyset} m(B)
\end{equation}
\textbf{Proof: BPA Construction}
Each tree represents an independent evidence source. The BPA is the proportion of trees supporting the union of intervals in $A$, satisfying:
\begin{itemize}
    \item $m(\emptyset) = 0$ (impossible event)
    \item $\sum_{A\subseteq\Omega} m(A) = 1$ (normalization)
\end{itemize}
The finite frame ensures all sums are well-defined and finite.
\subsubsection*{Belief Function:}
For nested intervals $A_1 \subseteq A_2 \subseteq \cdots \subseteq A_n$ defined over the real line, we map to our finite frame by taking $A_{j,\Omega} = \{I \in \Omega : I \subseteq A_j\}$. Then:
\begin{equation}
\text{Bel}(A_n) = \sum_{B \subseteq A_{n,\Omega}} m(B) \quad \text{(consonant structure)}
\end{equation}
This follows from the definition of Belief as the total evidence supporting all subsets of $A_{n,\Omega}$.
\subsubsection*{Plausibility Bound:}
For conflicting explanations (e.g., positive vs. negative impact), let $A$ be an interval and $A^c$ its complement. The ambiguity is:
\begin{equation}
\text{Pl}(A) - \text{Bel}(A) = 1 - \sum_{B \subseteq A_\Omega} m(B) - \sum_{B \subseteq A^c_\Omega} m(B)
\end{equation}
where $A_\Omega = \{I \in \Omega : I \subseteq A\}$ and $A^c_\Omega = \{I \in \Omega : I \subseteq A^c\}$. This quantifies the probability mass assigned to sets overlapping both $A$ and $A^c$.
\subsubsection*{Tree Ensemble Specialization:}
Since trees are exchangeable and the frame is finite:
\begin{equation}
\lim_{K \to \infty} \text{Bel}(A) = \mathbb{P}(\phi_i \in A)
\end{equation}
By the Law of Large Numbers, Belief converges to the true probability when the partition is sufficiently refined.
\\
\\
\textbf{Conflict Measure:} The explanation conflict for feature $i$ is:
\begin{equation}
\mathcal{C}_i = \max_{j=1,\dots,M} \left[ \text{Pl}(I_j) - \text{Bel}(I_j) \right]
\end{equation}
which measures the maximum ambiguity in SHAP assignments across the partition intervals.
\subsubsection*{Theoretical Implications}
This finite-frame Dempster-Shafer representation enables rigorous uncertainty quantification for SHAP values in tree ensembles:
\begin{enumerate}
    \item The finite frame ensures all belief and plausibility sums are finite and well-defined.
    \item The partition-based approach aligns with histogram estimation of SHAP distributions.
    \item Belief represents confirmed evidence while plausibility represents possible evidence, enabling precise uncertainty decomposition.
    \item The conflict measure $\mathcal{C}_i$ identifies features with contradictory importance patterns across trees.
\end{enumerate}
For high-stakes AI systems, this evidence-theoretic framework provides a mathematically sound foundation for reliability assessment of feature importance explanations, complementing the variance-based decomposition in Equation~\eqref{eq:shap_decomposition}.

\subsection*{Uncertainty Theory: Application to SHAP}
Uncertainty theory by Liu et al.~\cite{liu2013toward, liu2024modified} is a mathematical framework designed to address epistemic uncertainty arising from incomplete knowledge, small sample sizes, or reliance on expert judgment. The theory is based on four axioms: normality, monotonicity, self-duality, and countable subadditivity. The central concept is the uncertainty distribution, denoted by the symbol $\Gamma \colon \mathbb{R} \to [0,1]$, which quantifies the degree of belief that a variable takes on values less than or equal to $\phi_i$.
This distribution is characterized by a value of 0 for implausible values and a value of 1 for fully plausible values. 
It is also monotonically increasing as values become more plausible. 
Uncertainty distributions model subjective confidence rather than frequency or likelihood. 
This makes them useful in situations with limited or non-statistical data. When applied to SHAP or feature attribution in AI, an uncertainty distribution can represent confidence in a feature's attribution magnitude. 
This allows practitioners to explicitly model and quantify their uncertainty about the importance of each feature, especially when data is scarce or unreliable. 
Entropy minimization can guide optimal data acquisition, thereby improving model interpretability and reliability~\cite{lio2018residual, liu2013toward, liu2024modified, zhou2023systematic, najafi2022fractional}.

\begin{theorem}[Uncertainty Distribution]
The uncertainty distribution $\Gamma: \mathbb{R} \to [0,1]$ for SHAP value $\phi_i$ satisfies:
\begin{enumerate}
    \item $\Gamma(c) = 0$ for $c < \min_k \phi_i^{(k)}$
    \item $\Gamma(c) = 1$ for $c \geq \max_k \phi_i^{(k)}$
    \item $\Gamma$ is monotonically increasing
\end{enumerate}
\end{theorem}

\subsubsection*{Boundary Conditions:} 
By definition, implausible values (outside $[\min\phi, \max\phi]$) have $\Gamma(c) = 0$, and fully plausible values ($c \geq \max\phi$) have $\Gamma(c) = 1$.

\subsubsection*{Monotonicity:} 
For any $c_1 < c_2$:
\begin{equation}
\{ k : \phi_i^{(k)} \leq c_1 \} \subseteq \{ k : \phi_i^{(k)} \leq c_2 \}
\end{equation}
Thus $\Gamma(c_1) \leq \Gamma(c_2)$ by set inclusion.

\subsubsection*{Entropy Minimization:} 
The uncertainty entropy is:
\begin{equation}
\label{eq:uncertainity_entropy}
H(\Gamma) = - \int_{-\infty}^{\infty} \gamma(c) \log \gamma(c)  dc
\end{equation}
where $\gamma(c) = d\Gamma/dc$. Data acquisition minimizes $H(\Gamma)$ by:
\begin{equation}
\label{eq:uncertainty_data_acquisition_minimization}
\mathbf{x}^* = \arg\min_{\mathbf{x}} \mathbb{E}_{y|\mathbf{x}} \left[ H(\Gamma_{\mathcal{D} \cup (\mathbf{x},y)}) \right]
\end{equation}
This follows from the information gain principle.

\begin{lemma}[Optimal Acquisition]
When acquiring data for feature $j$, the uncertainty entropy decreases as:
\begin{equation}
\Delta H \propto -\text{Cov}\left( \phi_j, \frac{\partial \phi_i}{\partial \theta} \right)
\end{equation}
where $\theta$ is the model parameter space.
\end{lemma}

\subsection*{Dirichlet Process Hypothesis Sampling}

Dirichlet processes (DPs) are key to Bayesian nonparametric modeling. 
They allow for flexible inference over distributions with unknown and potentially infinite underlying clusters. DPs are useful for modeling uncertainty in complex spaces like SHAP values and their clusters across tree ensembles. 
Each sample from a DP is a discrete probability distribution. DPs are parameterized by a concentration parameter, $\alpha$, and a base distribution, $G_0$. 
In mixture models, DPs allow for an unbounded number of mixture components, thereby adapting model complexity to the data. 
In tree ensembles, each tree is a hypothesis about feature attributions. 
By modeling the distribution of SHAP values across trees with a DP mixture model, one can cluster SHAP values without specifying the number of clusters or modes in advance. 
This method captures both diversity and epistemic uncertainty in feature attributions due to model variability and quantifies uncertainty by examining the posterior distribution of clusters or modes of SHAP values. 
This provides richer uncertainty estimates than standard bootstrap or ensemble variance methods. 
DPs have several advantages over parametric and bootstrap methods. First, they avoid the need to fix the number of clusters or modes in advance. DPs adapt to model complexity as more data or trees are considered. 
DPs mitigate under- or overfitting and provide a more nuanced, probabilistic view of uncertainty in SHAP attributions. 
DPs are widely used in machine learning for clustering and mixture models where the number of components is unknown. 
Such an approach allows for a richer and more flexible quantification of epistemic uncertainty in SHAP attributions by leveraging the full power of Bayesian nonparametrics~\cite{li2019tutorial, lin2016dirichlet, reisberger2024linkage, tresp2006dirichlet, teh2017dirichlet, ng2011dirichlet}.

\begin{theorem}
[Constructing the Dirichlet Process]
The posterior over tree ensembles is given by:
\begin{equation}
\label{eq:dirichlet_tree}
G \sim \text{DP}(\alpha, G_0), \quad G_0 = \sum_{k=1}^K w_k \delta_{T_k}, \quad w_k = \frac{\text{OOB-AUC}_k}{\sum_j \text{OOB-AUC}_j}
\end{equation}
\end{theorem}

\subsubsection*{Base Measure:} 
$G_0$ is a discrete measure weighted by out-of-bag (OOB) accuracy, satisfying $\int dG_0 = 1$.

\subsubsection*{Dirichlet Process:} 
For any partition $(B_1,\dots,B_m)$ of the tree space:
\begin{equation}
(G(B_1),\dots,G(B_m)) \sim \text{Dirichlet}(\alpha G_0(B_1), \dots, \alpha G_0(B_m))
\end{equation}

\subsubsection*{Concentration Parameter:}
\begin{itemize}
    \item As $\alpha \to 0$: $G$ concentrates on $\max(w_k)$ trees
    \item As $\alpha \to \infty$: $G \to G_0$ (base measure)
\end{itemize}

\subsection*{SHAP Distribution:} 
The SHAP value distribution is:
\begin{equation}
\label{eq:shap_distribution_dirichlet}
\mathbb{F}_i(A) = \int \phi_i(T)  dG(T)
\end{equation}
With first moment:
\begin{equation}
\mathbb{E}[\phi_i] = \sum_{k=1}^K \pi_k \phi_i^{(k)}, \quad \pi \sim \text{Dirichlet}(\alpha \mathbf{w})
\end{equation}

\begin{theorem}[Convergence]
As $K \to \infty$, the SHAP distribution converges:
\begin{equation}
\label{eq:convergence_gaussian}
\mathbb{F}_i \xrightarrow{d} \mathcal{GP}\left( m(\mathbf{x}), \kappa(\mathbf{x},\mathbf{x}') \right)
\end{equation}
where $m(\cdot)$ is the mean function and $\kappa(\cdot,\cdot)$ the covariance kernel.
\end{theorem}

\begin{enumerate}
    \item By the de Finetti theorem, infinite exchangeable trees induce a Gaussian process~\cite{freer2009computable, freer2010posterior, freer2012computable, royprobabilistic}.
    \item The Dirichlet process is the de Finetti measure for Pólya sequences~\cite{mauldin1992polya}.
    \item SHAP values are continuous linear operators, preserving convergence~\cite{lundberg2017unified}.
\end{enumerate}

\section*{METHODS}
\label{sec:methodology}

Our approach integrates three complementary theoretical frameworks to facilitate the decomposition and quantification of uncertainty in SHAP values: Dirichlet process hypothesis sampling, Liu's uncertainty theory, and Dempster–Shafer theory.  
This integrated approach explicitly models the entanglement between aleatoric and epistemic uncertainties in feature attribution, overcoming the drawbacks of traditional uncertainty quantification. 
The framework allows for a thorough examination of sources of uncertainty(Algorithm:~\ref{alg:etreefleccomb}).

\subsubsection*{Evidence Theory for SHAP Uncertainty}
Dempster-Shafer evidence theory provides a formal mechanism to represent ambiguity in SHAP distributions through belief functions. For a tree ensemble with $K$ trees, we construct a Basic Probability Assignment (BPA) over SHAP intervals $A \subseteq \mathbb{R}$ (Equation:~\ref{eq:m(a)_DST}).

where $\phi_i^{(k)}$ denotes the SHAP value from tree $T_k$ (Algorithm:~\ref{alg:constrained}). The belief $\text{Bel}(A)$ and plausibility $\text{Pl}(A)$ functions then quantify the minimum and maximum support for interval $A$, respectively. The physical interpretation reveals that $\text{Bel}(A)$ represents conservative certainty (e.g., "SHAP lies in $[-1,1]$ with $\geq$80\% confidence"), while the conflict measure $\mathcal{C}_i = \sup_A [\text{Pl}(A) - \text{Bel}(A)]$ captures explanation ambiguity. High conflict triggers human verification in critical applications, and the BPA dispersion directly measures aleatoric uncertainty. This approach links to the SHAP variance decomposition by mapping belief/plausibility bounds to epistemic uncertainty ($\text{Var}_f(\mathbb{E}_D[\phi_i|f])$) and the conflict term to entanglement ($\mathcal{C}(f,D)$).

\subsubsection*{Uncertainty Theory for SHAP Uncertainty}
Liu et al. uncertainty theory models epistemic uncertainty through the uncertainty distribution $\Gamma(c) = \mathbb{P}(\phi_i \leq c)$, bounded by $[\min_k \phi_i^{(k)}, \max_k \phi_i^{(k)}]$. The distribution's shape provides intuitive study of the uncertainty : a steep $\Gamma$ indicates low epistemic uncertainty (tight SHAP concentration), while a flat $\Gamma$ reflects high epistemic uncertainty (broad dispersion). The median SHAP value occurs at $\Gamma(c)=0.5$. We operationalize this framework through uncertainty entropy minimization (Equation:~\ref{eq:uncertainity_entropy}).

which guides optimal data acquisition (Equation:~\ref{eq:uncertainty_data_acquisition_minimization}).

This entropy reduction disproportionately targets features with high $\text{Var}(\phi_j)$, thereby reducing aleatoric uncertainty ($\mathbb{E}_D[\text{Var}_f(\phi_i|D)]$). The theory explicitly quantifies epistemic uncertainty through $\Gamma$'s spread, complementing the evidence theory framework (Algorithm:~\ref{alg:aggregation}).

\subsubsection*{Dirichlet Process Hypothesis Sampling}
Dirichlet process (DP) hypothesis sampling (Algorithm:~\ref{alg:dirichlet}) integrates both aleatoric and epistemic uncertainty through Bayesian nonparametrics. We model the posterior over tree ensembles (Equation: ~\ref{eq:dirichlet_tree}).

where $G_0$ weights trees by out-of-bag reliability. The concentration parameter $\alpha$ controls uncertainty estimation: $\alpha \ll 1$ focuses on high-accuracy trees (low epistemic uncertainty), while $\alpha \gg 1$ enforces uniform weighting (high epistemic uncertainty). SHAP distributions are derived as $\mathbb{F}_i(A)$ (Equation:~\ref{eq:shap_distribution_dirichlet}).

As $K \to \infty$, $\mathbb{F}_i$ converges to a Gaussian process (Equation:~\ref{eq:convergence_gaussian}), preserving SHAP linearity. This method captures aleatoric uncertainty through within-sample SHAP variance and epistemic uncertainty through between-sample variance of $\mathbb{E}[\phi_i]$, while maintaining entanglement via the DP's covariance structure.

The three frameworks form an end-to-end workflow that decomposes SHAP variance (Algorithm:~\ref{alg:decomposition}) (Equation:~\ref{eq:shap_decomposition}). Evidence theory quantifies epistemic uncertainty and conflict, Liu's theory models epistemic spread and guides data acquisition, and Dirichlet sampling integrates both through its weighted nonparametric formulation. The interconnection manifests in three key linkages: (1) Conflict detection (evidence theory) flags features for entropy minimization (Liu's theory); (2) Dirichlet samples generate distributions feeding into $\text{Bel}/\text{Pl}$ and $\Gamma$ calculations; (3) Data acquisition refines $G_0$ in the DP base measure. This triad addresses the SHAP uncertainty decomposition as follows: aleatoric uncertainty is measured through BPA dispersion (evidence theory) and within-DP-sample variance; epistemic uncertainty is quantified by $\text{Pl}(A) - \text{Bel}(A)$, $\Gamma$-entropy, and $\alpha$-driven hypothesis sampling; entanglement is preserved via conflict terms $\mathcal{C}_i$ and the DP's covariance structure. The unified methodology (Algorithm:~\ref{alg:etreefleccomb}) enables granular attribution of uncertainty sources critical for high-stakes interpretability.
\subsection*{Physical Interpretation}

\subsubsection*{Evidence Theory}
\begin{itemize}
    \item Belief: Minimum support for SHAP interval
    \item Plausibility: Maximum possible support
    \item Conflict: $\text{Pl}(A) - \text{Bel}(A) > 0$ indicates ambiguous explanations
\end{itemize}

\subsubsection*{Uncertainty Distribution}
\begin{itemize}
    \item $\Gamma(c) = 0.5$ at median SHAP value
    \item Steep $\Gamma$ $\Rightarrow$ low epistemic uncertainty
    \item Flat $\Gamma$ $\Rightarrow$ high epistemic uncertainty
\end{itemize}

\subsubsection*{Dirichlet Process}
\begin{itemize}
    \item $\alpha$ controls ``exploration-exploitation'' of hypothesis space
    \item $w_k$ weights represent tree reliability
    \item Samples $G$ represent plausible realizations of the model
\end{itemize}

\subsection*{Practical Implications}

\begin{itemize}
    \item Conflicting Explanations: High $\text{Pl}(A) - \text{Bel}(A)$ triggers human verification in critical applications.
    \item Data Acquisition: Minimizing $H(\Gamma)$ focuses data collection on high-uncertainty features:
                            \begin{equation}
                            \frac{\partial H}{\partial n_j} \propto -\text{Var}(\phi_j)
                            \end{equation}
    \item Hypothesis Sampling: The Dirichlet concentration parameter $\alpha$ controls uncertainty estimation:
                                \begin{itemize}
                                    \item $\alpha \approx 1$: Balanced exploration
                                    \item $\alpha < 1$: Focus on best-performing trees
                                    \item $\alpha > 1$: Uniform uncertainty estimation
                                \end{itemize}
\end{itemize}

\begin{algorithm}
\caption{Dirichlet-Weighted Tree Sampling}
\label{alg:dirichlet}
\textbf{Purpose}: Generate hypothesis-consistent sub-ensembles \\
\textbf{Input}: Trained ensemble $\mathcal{M}$, training data $D$, concentration $\alpha$, temperature $\beta$ \\
\textbf{Output}: List of $S$ sub-ensembles

\begin{algorithmic}[1]
\Function{DirichletSample}{$\mathcal{M}$, $D$, $S$, $\alpha$, $\beta$}
    \For{each tree $T_k$ in $\mathcal{M}$}
        \State Compute OOB accuracy $a_k$ using $D$ 
        \State $w_k \gets \exp(\beta \cdot a_k) / \sum_j \exp(\beta \cdot a_j)$ \Comment{Softmax weighting}
    \EndFor
    \For{$s = 1$ to $S$}
        \State Draw $\pi \sim \text{Dirichlet}(\alpha \cdot w)$ \Comment{Dirichlet distribution}
        \State Sample tree indices $I \sim \text{Categorical}(\pi)$
        \State Construct sub-ensemble $\mathcal{M}_s = \{T_i \mid i \in I\}$
        \State \Return $\mathcal{M}_s$
    \EndFor
\EndFunction
\end{algorithmic}
\end{algorithm}

\begin{algorithm}
\caption{Constrained TreeSHAP Computation}
\label{alg:constrained}
\textbf{Purpose}: Compute SHAP values preserving path dependencies \\
\textbf{Input}: Sub-ensemble $\mathcal{M}_s$, instance $\mathbf{x}$, background data $B$ \\
\textbf{Output}: SHAP vector $\phi$

\begin{algorithmic}[1]
\Function{ConstrainedTreeSHAP}{$\mathcal{M}_s$, $\mathbf{x}$, $B$}
    \For{each tree $T$ in $\mathcal{M}_s$}
        \State $\phi_T \gets \text{TreeSHAP}(T, \mathbf{x}, B)$ \Comment{Standard TreeSHAP computation}
    \EndFor
    \State $\phi_{\text{mean}} \gets \text{mean}(\phi_T \text{ across trees})$
    \State $\Sigma \gets \text{Covariance}(\phi_T \text{ across trees})$ \Comment{Feature covariance matrix}
    \State $\phi_{\text{adj}} \gets \phi_{\text{mean}} + 0.5 \cdot \text{diag}(\Sigma)$ \Comment{Interaction adjustment}
    \State \Return $\phi_{\text{adj}}$
\EndFunction
\end{algorithmic}
\end{algorithm}

\begin{algorithm}
\caption{SHAP Variance Decomposition}
\label{alg:decomposition}
\textbf{Purpose}: Quantify uncertainty components \\
\textbf{Input}: SHAP distributions $\{\Phi_s\}_{s=1}^S$ \\
\textbf{Output}: Aleatoric, epistemic, entanglement terms

\begin{algorithmic}[1]
\Function{DecomposeVariance}{$\{\Phi_s\}$}
    \For{each feature $i$}
        \State $\mu_s[i] \gets \text{mean}(\Phi_s^i)$ \Comment{Within-sample mean}
        \State $\sigma^2_s[i] \gets \text{variance}(\Phi_s^i)$ \Comment{Within-sample variance}
        \State $A[i] \gets \text{mean}(\sigma^2_s[i])$ \Comment{Aleatoric uncertainty}
        \State $E[i] \gets \text{variance}(\mu_s[i])$ \Comment{Epistemic uncertainty}
        \State $C[i] \gets \text{Covariance}(\mu_s[i], \sigma^2_s[i])$ \Comment{Entanglement term}
    \EndFor
    \State \Return $(A, E, C)$
\EndFunction
\end{algorithmic}
\end{algorithm}

\begin{algorithm}
\caption{Uncertainty-Aware SHAP Aggregation}
\label{alg:aggregation}
\textbf{Purpose}: Compute final SHAP values with uncertainty metrics \\
\textbf{Input}: SHAP distributions $\{\Phi_s\}$, features $F$ \\
\textbf{Output}: Mean SHAP, uncertainty metrics
\begin{algorithmic}[1]
\Function{AggregateUncertainty}{$\{\Phi_s\}$, $F$}
    \For{each feature $i$ in $F$}
        \State $\mu[i] \gets \text{mean}(\Phi_s^i)$ \Comment{Mean SHAP value}
        \State $\sigma[i] \gets \text{std}(\Phi_s^i)$ \Comment{Standard deviation}
        \State $\text{CI}[i] \gets [\text{percentile}(\Phi_s^i, 2.5), \text{percentile}(\Phi_s^i, 97.5)]$ \Comment{95\% CI}
        \State $H[i] \gets \text{Entropy}(\Phi_s^i)$ \Comment{Differential entropy}
        \State $\text{SS}[i] \gets P(\text{sign}(\phi) \text{ constant})$ \Comment{Sign stability}
    \EndFor
    \State \Return $(\mu, \sigma, \text{CI}, H, \text{SS})$
\EndFunction
\end{algorithmic}
\end{algorithm}

\begin{algorithm}
\caption{UbiQTree End-to-End}
\label{alg:etreefleccomb}
\textbf{Purpose}: Full uncertainty quantification pipeline \\
\textbf{Input}: Model $\mathcal{M}$, data $D$, instance $\mathbf{x}$, parameters \\
\textbf{Output}: SHAP values with uncertainty

\begin{algorithmic}[1]
\Function{E\_SHAP}{$\mathcal{M}$, $D$, $\mathbf{x}$, $S=500$, $\alpha=0.5$, $\beta=5.0$}
    \State \Comment{Step 1: Hypothesis sampling}
    \State $\mathcal{M}_{\text{list}} \gets \text{DirichletSample}(\mathcal{M}, D, S, \alpha, \beta)$
    \State
    \State \Comment{Step 2: SHAP computation}
    \For{each $\mathcal{M}_s$ in $\mathcal{M}_{\text{list}}$}
        \State $\Phi_s \gets \text{ConstrainedTreeSHAP}(\mathcal{M}_s, \mathbf{x}, D)$
        \State Store $\Phi_s$
    \EndFor
    \State
    \State \Comment{Step 3: Variance decomposition}
    \State $(A, E, C) \gets \text{DecomposeVariance}(\{\Phi_s\})$
    \State
    \State \Comment{Step 4: Uncertainty metrics}
    \State $(\mu, \sigma, \text{CI}, H, \text{SS}) \gets \text{AggregateUncertainty}(\{\Phi_s\})$
    \State
    \State \Return \texttt{mean\_shap: $\mu$, std\_dev: $\sigma$, ci\_95: CI, aleatoric: A, epistemic: E, entanglement: C, entropy: H, sign\_stability: SS}
\EndFunction
\end{algorithmic}
\end{algorithm}

\section*{RESULTS}
To study and analyze epistemic uncertainty in these experiments, we implemented our framework. We performed an ensemble-based SHAP analysis for each class in our dataset. Then, we plotted the mean absolute SHAP value for each class in our dataset using the trained model. For this study, we trained an RF classifier across various datasets. In this study, we relied on the absolute SHAP values for each class in the dataset. These values are useful for comparing the relative strength of a feature's contribution within each class, measuring the uncertainty of a feature's impact on the class's output, and identifying features that the model considers decisive for a class, regardless of whether they increase or decrease the class's logit/probability. Absolute SHAP is particularly well-suited for quantifying uncertainty per class because it allows us to analyze the stability of a feature's influence on a given class. Furthermore, it allows us to evaluate whether the model consistently demonstrates confidence in the feature's significance for the class, regardless of its sign. SHAP variance or entropy indicates the robustness of class-specific attribution magnitude across model variants. The $\pm2\sigma$  (standard deviation) is plotted on the mean absolute SHAP chart. This $\pm2\sigma$  denotes epistemic uncertainty. Features are then ranked by mean contribution. Relatively wide violin plot indicate considerable variability across different instantiations of sub-ensembles. Higher contributions show that the model consistently relies on this feature. Along with the wide violin plot, they indicate that its impact magnitude is not well understood and that there is a lot of uncertainty about it. The narrow violin plot on the chart represent high-confidence features that contribute to stable predictions. We select the top three features with the highest contributions from each class of the different datasets; however, the user can select as many as required and analyze them. SHAP distribution analysis is performed to evaluate the stability of features that contribute the most across model instances. A high standard deviation suggests that the SHAP values are inconsistent across subtrees or subensembles, indicating uncertainty about each feature's influence. The SHAP distribution analysis also reveals explanation entropy. High entropy indicates a flat or dispersed distribution of SHAP values, indicating low certainty about the features' impact. Consistent, peaked SHAP value distributions indicate low entropy. The SHAP distribution visualization helps users understand the directional stability of SHAP values. This measure quantifies the consistency of the sign of the SHAP value, providing insight into whether the SHAP values contribute negatively or positively across sub-models. We group features into three categories based on their directional consistency: high stability (>=90\%), moderate stability (>=67\%), and low stability (<67\%). This allows us to study how the interpretation values vary despite having varying epistemic uncertainty. This provides insight into the validity of the SHAP values in any given model realization. Overall, quantifying epistemic uncertainty and visualizing SHAP magnitude distributions, distributional shapes, entropies, and directional consistencies helps us quantify uncertainty in terms of feature contributions or importance, as calculated using SHAP values. Analyzing the subtrees in ensemble methods helps us simulate posterior samples from the model space. This allows us to focus on uncertainty in the model rather than data noise. We control our methodology using the number of posterior samples, or the $\alpha$ parameter. We select different parameters for each dataset to validate our approach and collect insights. By extracting and explaining predictions using different subsets of trees (or different models in an ensemble), we simulate how the model would behave under slightly different yet still plausible versions of itself. This captures variability due to uncertainty in model specification. SHAP explanations reveal differences via sub-models and show how much confidence can be placed in a specific feature attribution. This makes them sensitive to the model's structure. The epistemic uncertainty thresholds are chosen based on experiments: 0.05 <= $\sigma$ < 0.1 requires expert-in-the-loop verification; 0.05 <= $\sigma$ < 0.05 is used for automated decisions; and $\sigma$ >= 0.1 suggests model retraining. 
A mean $\pm2\sigma$ SHAP chart shows a feature's average effect and variability. Wide variance across submodels implies high uncertainty. 
A SHAP kernel density estimate (KDE) along with the confidence interval (CI) plot visualizes the distribution of SHAP values. 
Multimodality or flatness indicates disagreement between model variants. Finally, quantitative uncertainty metrics, such as SHAP distribution plot which indicates the standard distriution, entropy, and sign stability, offer summary statistics that allow us to directly assess the robustness and stability of a feature's contribution from the perspective of evidence theory.

\subsection*{Medical Information Mart for Intensive Care III (MIMIC-III) Dataset}

\begin{figure}[h!]
    \centering
    \includegraphics[width=\textwidth]{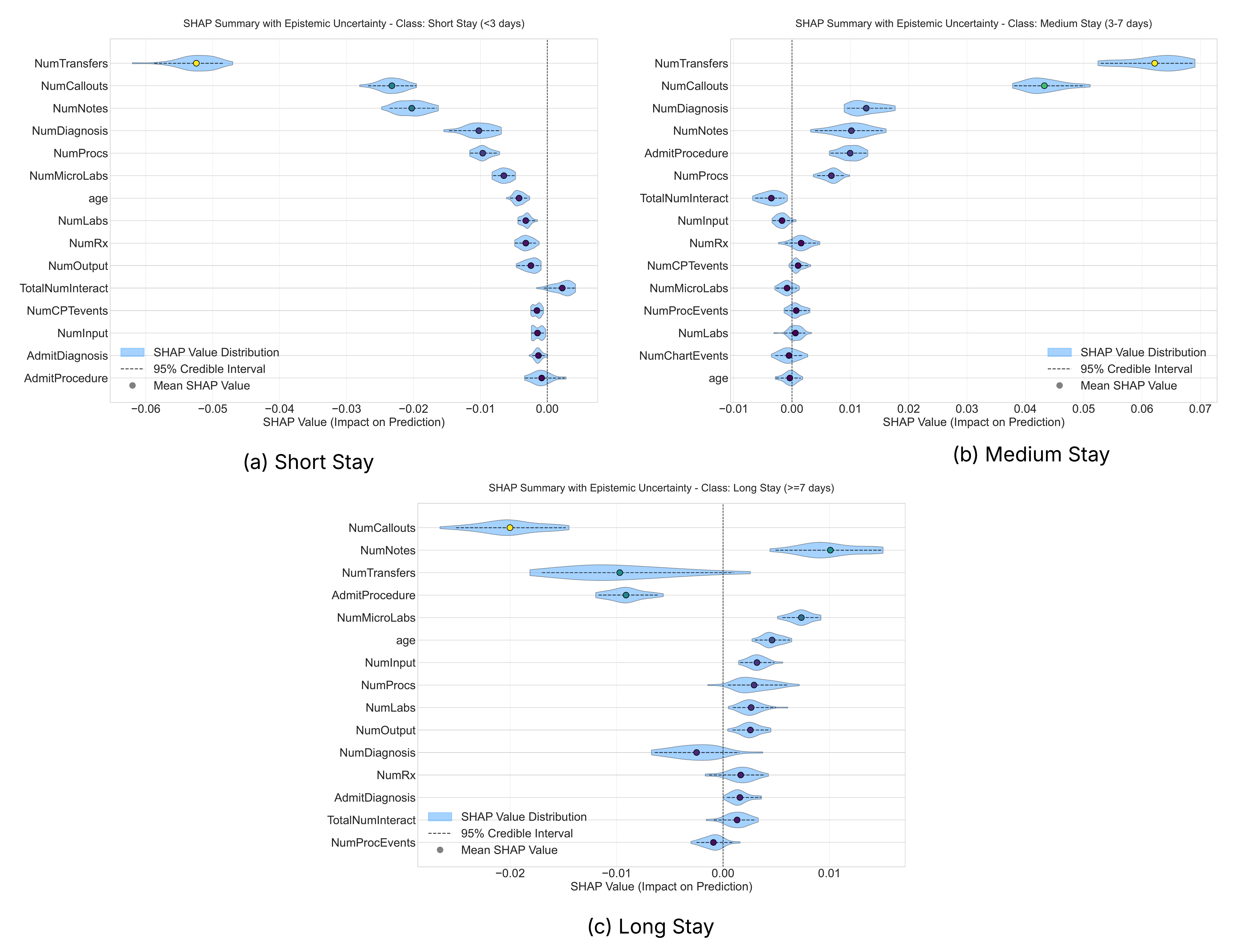}
    \caption{}
    \label{fig:mimic_los_shap}
\end{figure}
\begin{figure}[h!]
    \centering
    \includegraphics[width=\linewidth]{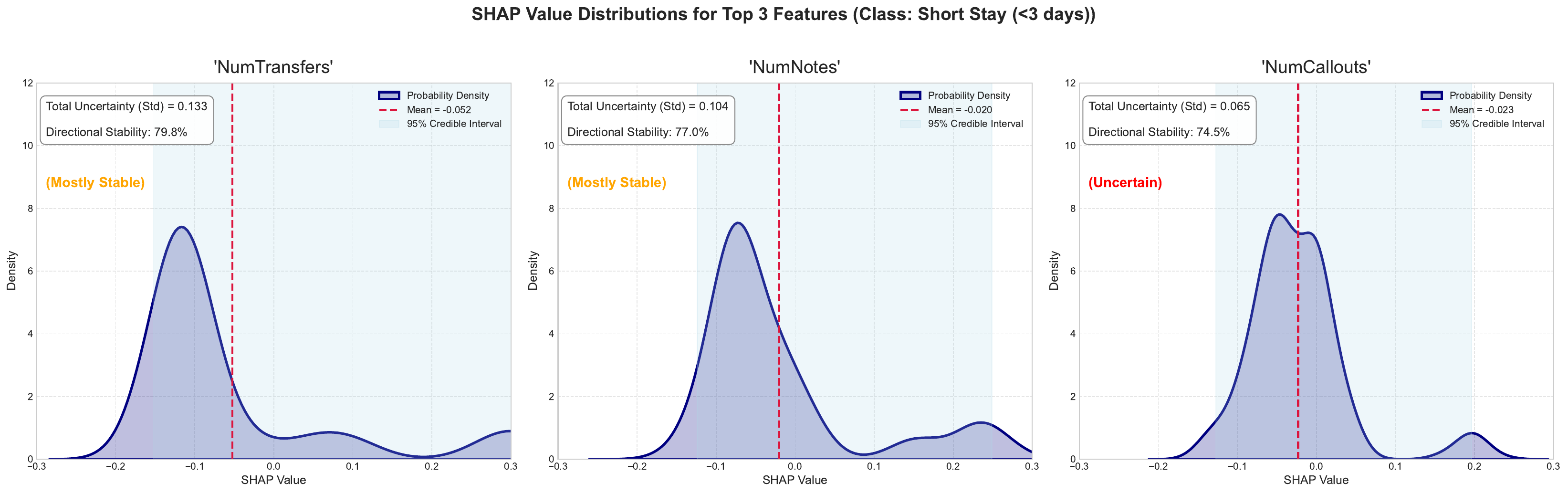}
    \caption{}
    \label{fig:mimic_los_nostay_shap_distribution}
\end{figure}

\begin{figure}[h!]
    \centering
    \includegraphics[width=\linewidth]{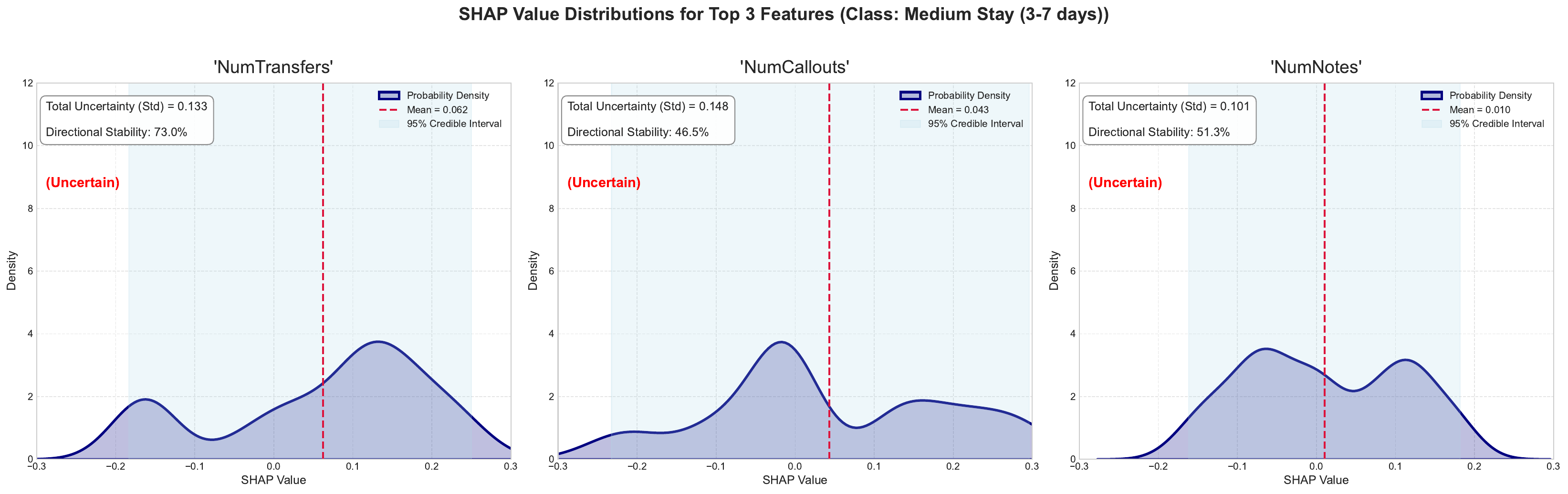} 
    \caption{}
    \label{fig:mimic_los_veryshort_shap_distribution}
\end{figure}

\begin{figure}[h!]
    \includegraphics[width=\linewidth]{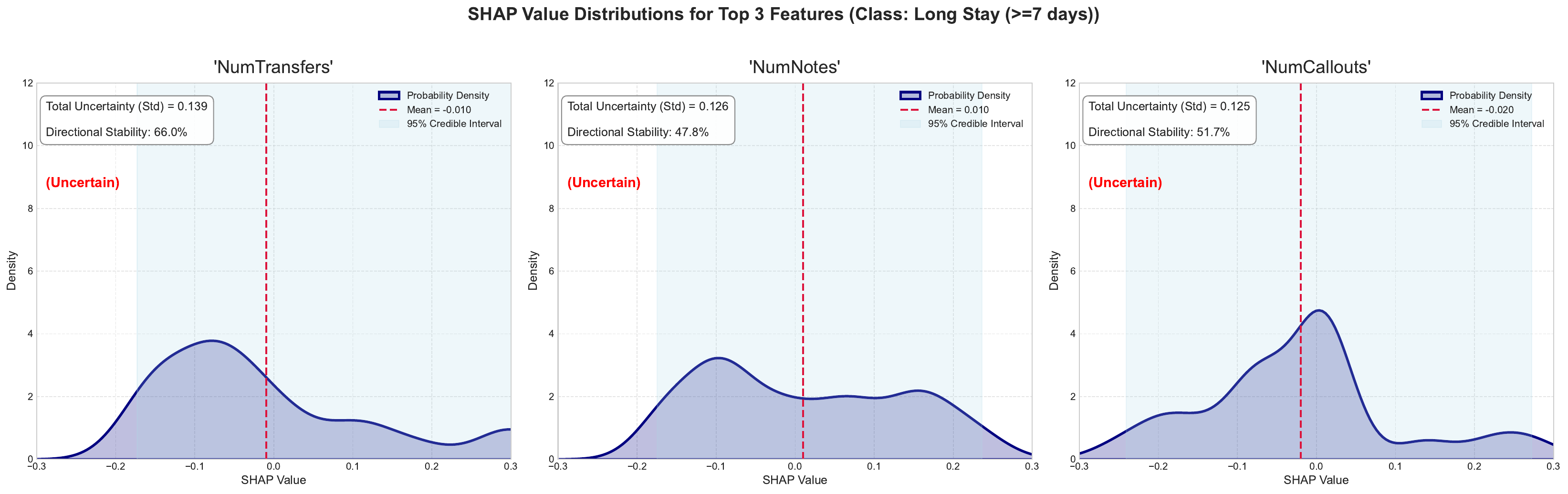}
    \caption{}
    \label{fig:mimic_los_short_shap_distribution}
\end{figure}

The Medical Information Mart for Intensive Care III (MIMIC-III)~\cite{johnson2016mimic} is a substantial clinical database comprising detailed health-related data from over 40,000 adult patients admitted to critical care units at a tertiary care hospital between 2001 and 2012. The dataset under consideration is extensive in nature, encompassing a wide range of information pertinent to the subject. This includes demographic data, vital signs, laboratory test results, medications, procedures, clinical notes, imaging reports, and hospital length of stay. The features \textit{'hadm\_id', 'LOSdays', 'religion', 'marital\_status', 'ethnicity'} were removed for ethical and privacy reasons and to replicate the real-world scenarios. We utilised this dataset to classify the datapoints into \textit{length of stay (LOS)}. We then, grouped the classes into \textit{Short Stay, Medium Stay, and Long Stay} categories. \textit{Short Stay} applies to patients who are assessed but not admitted to the hospital or refers to hospitalizations lasting less than three days, while \textit{Medium Stay} covers stays from three to seven days. \textit{Long Stay} describes hospitalizations exceeding seven days~\cite{zebin2019deep}. The feature descriptions for the features we have discussed are as follows, \textit{NumTransfers} represents the number of times a patient is transferred within the hospital during their stay. These transfers may occur between different units, such as from the emergency department to the ICU, or between wards. They reflect patient movement within the hospital. \textit{NumNotes} is the count of clinical notes recorded for a patient during their hospital admission or ICU stay. These notes include documentation from physicians, nurses, and other healthcare providers, capturing clinical observations, treatments, and progress. \textit{NumCallouts} is the count of recorded events when a patient was flagged for potential discharge or transfer from the ICU. 

\paragraph{For the reproducibility of these experiments, the following experimental details are provided: }
The MIMIC-III dataset was loaded, and non-informative columns (\textit{hadm\_id, religion, marital\_status, ethnicity}) were removed. The target variable, \textit{LOSgroupNum}, was obtained by categorizing the \textit{LOSdays} column into three groups: short stay (<3 days), medium stay (3-7 days), and long stay (>=7 days). Features were separated into numerical and categorical types. Median imputation was applied to numerical features to handle missing values, followed by standardization using \textit{StandardScaler}. The RF Classifier was trained with 100 estimators (\textit{n\_estimators=100}) with parameters \textit{class\_weight="balanced"} and \textit{oob\_score=True}. The data were prepared using a 70:30 train-test split with stratification. The model's performance was evaluated using the micro-averaged F1 score on the test set which was 94.70\%. Implementing the proposed framework on the test data assessed the epistemic uncertainty inherent in predictions for unseen data. The number of posterior samples drawn from the model space via a Dirichlet distribution (\textit{n\_dirichlet\_samples}) was set to 30. The Dirichlet distribution's concentration parameter, $\alpha$, was set to 0.5. When $\alpha$ < 1, sampling prioritizes sub-models (trees) with higher out-of-bag accuracy, focusing the analysis on the best-performing components of the ensemble. A comprehensive uncertainty analysis was performed for each of the three target classes \textit{Short Stay, Medium Stay, and Long Stay}. This included generating global feature importance plots (mean absolute SHAP and coefficient of variation), plotting SHAP value distributions with 95\% credible intervals for the top features, decomposing the uncertainty into epistemic, aleatoric, and entanglement components, and visualizing the results using the UbiQTree framework.(see Figure: ~\ref{fig:mimic_los_shap}). 

The SHAP summary plot provides a visual representation of the fifteen features having the highest feature importances and their impact on a model’s prediction for
the \textit{Short Stay, Medium Stay, \& Long Stay class}, incorporating epistemic uncertainty. To The violin plots illustrate the distribution of SHAP values for each feature. The blue shaded area indicates the 95\% confidence interval($\pm2\sigma$) of the feature’s impact on the prediction. This symbolizes the epistemic uncertainty present within the sub-ensembles and their 2$\sigma$ range across four distinct categories of the \textit{LOS}. In the case of the. In the context of the \textit{Short Stay} class, the top three features that exhibited the highest absolute SHAP values were identified as \textit{NumTransfers, NumNotes, NumCallouts}. For the class \textit{Medium Stay}, the three features that exhibit the highest absolute SHAP values are \textit{NumTransfers,
NumCallouts, NumNotes}. In the context of the \textit{Long Stay} class, the top three features with the highest absolute SHAP values are \textit{NumTransfers, NumNotes, NumCallouts}. The results from the absolute mean SHAP values chart (Figure:~\ref{fig:mimic_los_shap}) indicates that the SHAP feature importance in the underlying models varies by a high factor. It could also be noted that the features such as \textit{gender \& ExpiredHospital} don't have very high feature importance but still they have less variability. This indicates a considerable variability across different model sub-ensembles. This suggests that while the model consistently relies on the features such as \textit{NumTransfers, NumNotes, NumDiagnosis, \& AdmitProcedures} features, it does so with substantial epistemic ambiguity regarding its precise impact magnitude. In contrast, features such as \textit{NumLabs \& age} exhibit low average SHAP values and narrow uncertainty violin plots, reflecting both low importance and high confidence in their negligible contribution, indicating a stable but marginal roles in the prediction of the different classes. The one of the major differences that is observed is in terms of the positively or negatively influencing the predictions. 

Furthermore, we examined the top three features, their associated epistemic uncertainty, sign stability, and SHAP value distribution, as well as the mean SHAP and 95\% SHAP confidence interval. For the \textit{Short Stay} category, the epistemic uncertainties for the features \textit{NumTransfers, NumNotes, and NumCallouts} are 0.133, 0.104, and 0.065, respectively (Figure:~\ref{fig:mimic_los_nostay_shap_distribution}). The sign stability for these features is 79.8\%, 77.0\%, and 74.5\%, respectively. These metrics reflect model variance in attribution for each feature. The high standard deviation of epistemic uncertainties in the top three features suggests inconsistent SHAP values across the ensemble and reflects uncertainty about the features' influence. Explanation entropy for the features \textit{NumTransfers} is considerably high as well, indicated by a non-uniform distribution of SHAP values, signaling low information certainty about the feature's impact. The feature \textit{NumNotes} has high entropy, indicated by skewed SHAP value distributions. \textit{NumNotes} have sign stability of 77.0\%, indicating interpretive inconsistency and high epistemic uncertainty. For the \textit{Medium Stay} category, the epistemic uncertainties for the features \textit{NumTransfers, NumCallouts, and NumNotes} are 0.133, 0.148, and 0.101, respectively. The sign stabilities for these features are 73.0\%, 46.5\%, and 51.3\%, respectively (Figure~\ref{fig:mimic_los_veryshort_shap_distribution}). The high standard deviation of epistemic uncertainties in the top three features suggests inconsistent SHAP values across the ensemble and reflects uncertainty about the features' influence. The sign stabilities are low, indicating interpretive inconsistency with high epistemic uncertainty. For the \textit{Long Stay} category, the epistemic uncertainties for the features \textit{NumTransfers, NumNotes, and NumCallouts} are 0.139, 0.126, and 0.125, respectively. The sign stabilities for these features are 66.0\%, 47.8\%, and 51.7\%, respectively (Figure:~\ref{fig:mimic_los_short_shap_distribution}). The standard deviation of epistemic uncertainties is high in the top three features, suggesting inconsistent SHAP values across the ensemble and reflecting uncertainty about the features' influence. Explanation entropy for the features \textit{NumTransfers, NumNotes, and NumCallouts} is considerably high as well, indicated by a non-uniform distribution of SHAP values, signaling high information uncertainty about the feature's impact. The sign stabilities are unstable, indicating interpretive inconsistency and high epistemic uncertainty. 
\begin{figure}[h]
    \centering
    \includegraphics[width=\textwidth]{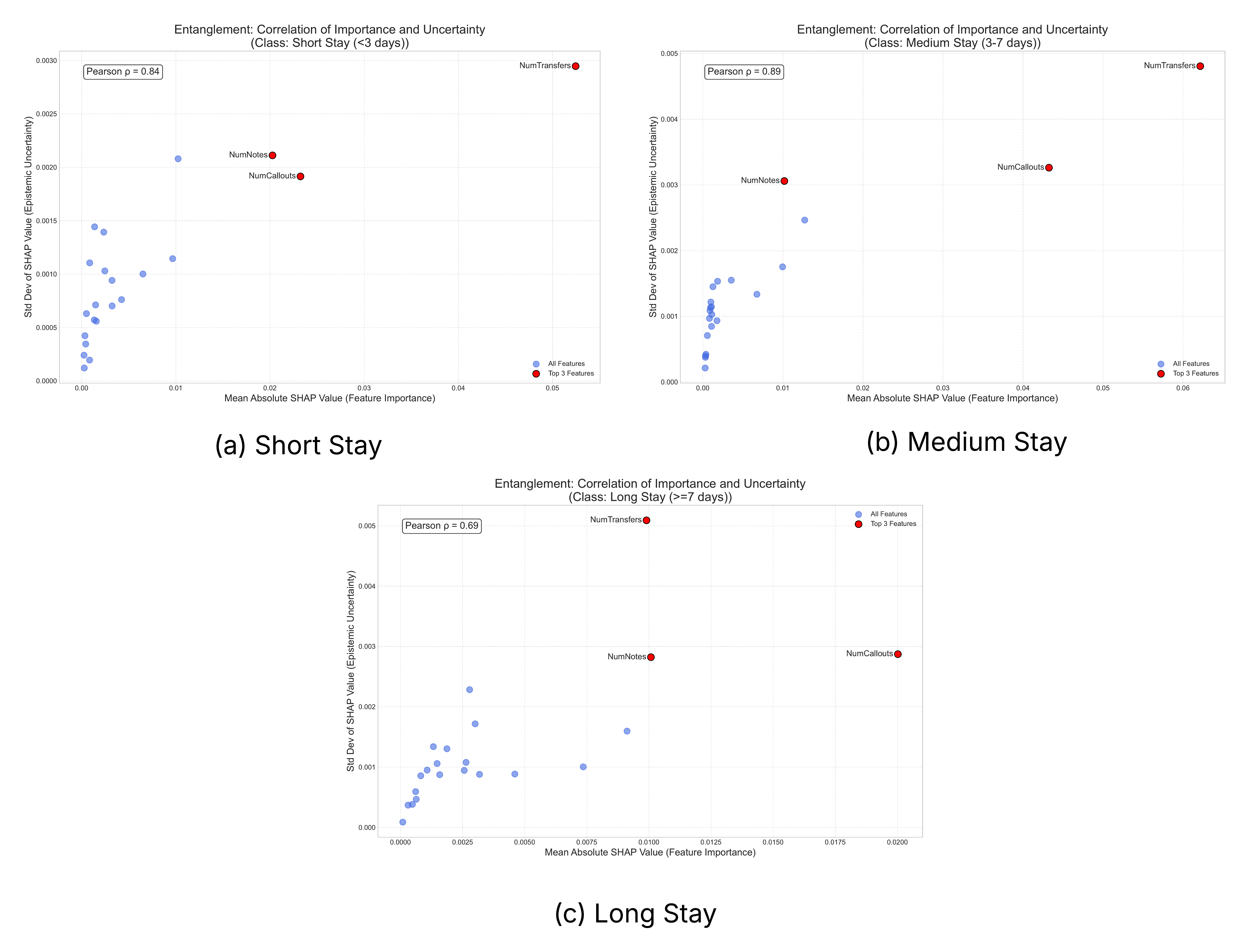}
    \caption{}
    \label{fig:mimic_entanglement}
\end{figure}
Further, a necessary investigation would be to study the entanglement term. The analysis revealed a strong positive correlation ($\rho$) between feature importance and explanation uncertainty across all patient length of stay (See Figure:~\ref{fig:mimic_entanglement}). The Pearson coefficients ($\rho$) were 0.84 for short stays, 0.89 for medium stays, and 0.69 for long stays. The most important features, \textit{NumTransfers, NumNotes, and NumCallouts}, are also the ones with the most unstable and unreliable SHAP values. This effect is most extreme for the medium-stay group, indicating that it is the most ambiguous and difficult-to-predict cohort. These findings imply a significant trust deficit in the model's explanations. The model's core reasoning for its predictions is highly sensitive to randomness in training, which makes the explanations volatile. To reliably deploy the model, especially for medium-stay predictions, it must be improved by stabilizing how these key features are processed to reduce epistemic uncertainty. In conclusions, an analysis of this dataset for hospital length-of-stay (LOS) reveals that the features the model uses most to make predictions i.e. \textit{NumTransfers, NumNotes, \& NumCallouts} are also its greatest sources of uncertainty and instability. This entanglement is strongest for the medium-stay (3–7 days) cohort, which is the most clinically ambiguous and difficult to predict. For decision-makers, this means that while the model identifies key drivers of patient stay duration, the reasoning behind classifying a patient as medium-stay is highly volatile and should not be trusted for high-stakes decisions such as resource planning or early discharge protocols.
The provided visualizations are important in achieving robust decision-making. SHAP summary plots with epistemic uncertainty flag features that are important yet unreliable, such as \textit{NumTransfers} for medium-stay patients. This allows domain professionals to question the model's logic and prioritize gathering more stable data or developing better features for these high-impact variables. SHAP distribution grids further reveal sign instability for features such as \textit{NumCallouts}, showing that their impact on predictions fluctuates between positive and negative across different model runs. This prevents dangerous overreliance on a single, seemingly definitive explanation.

\subsection*{Ovarian Cancer Dataset}
\begin{figure}[h!]
    \centering
    \includegraphics[width=\textwidth]{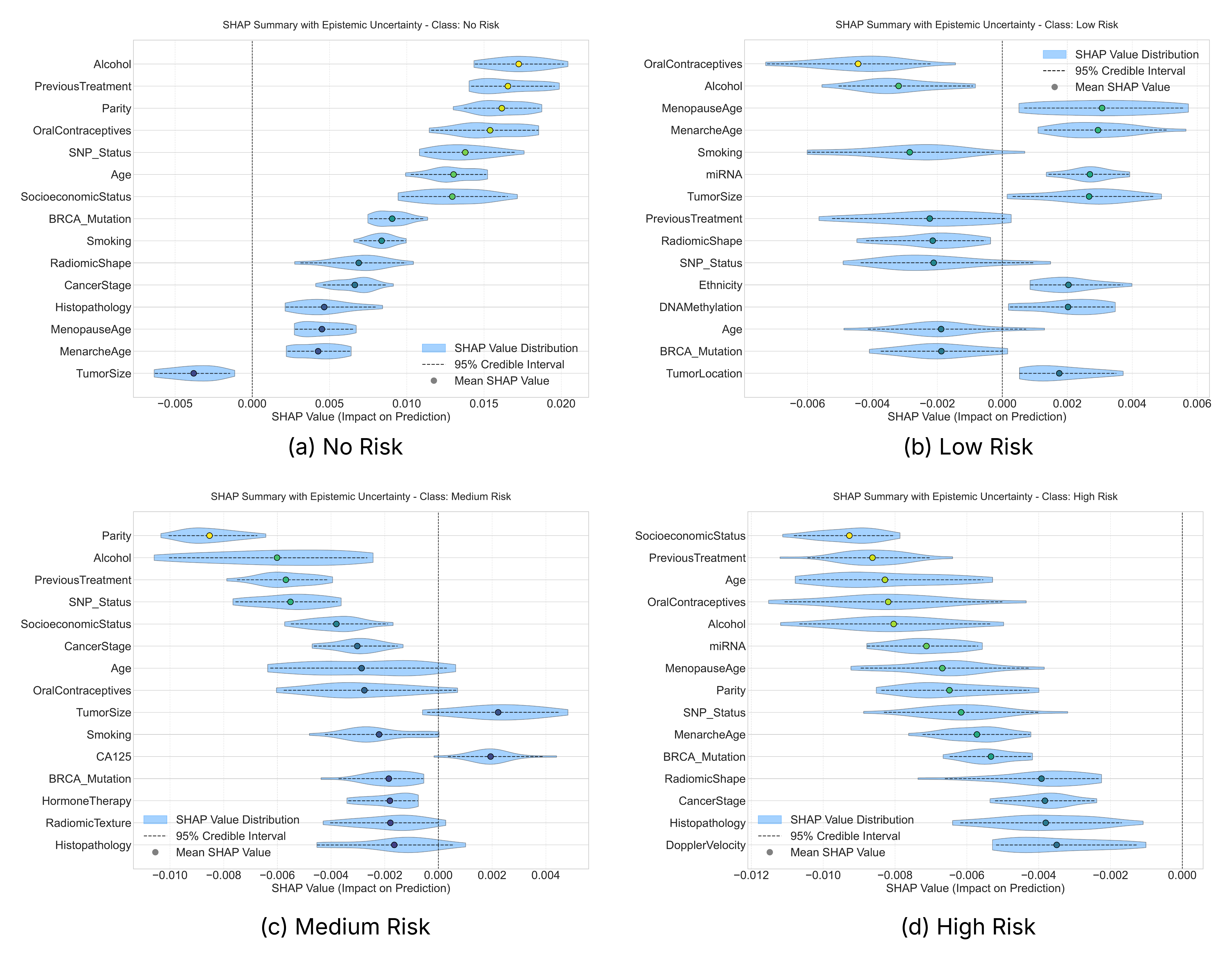}
    \caption{}
    \label{fig:ovarian_cancer_shap}
\end{figure}
\begin{figure}[h!]
    \centering
    \includegraphics[width=\linewidth]{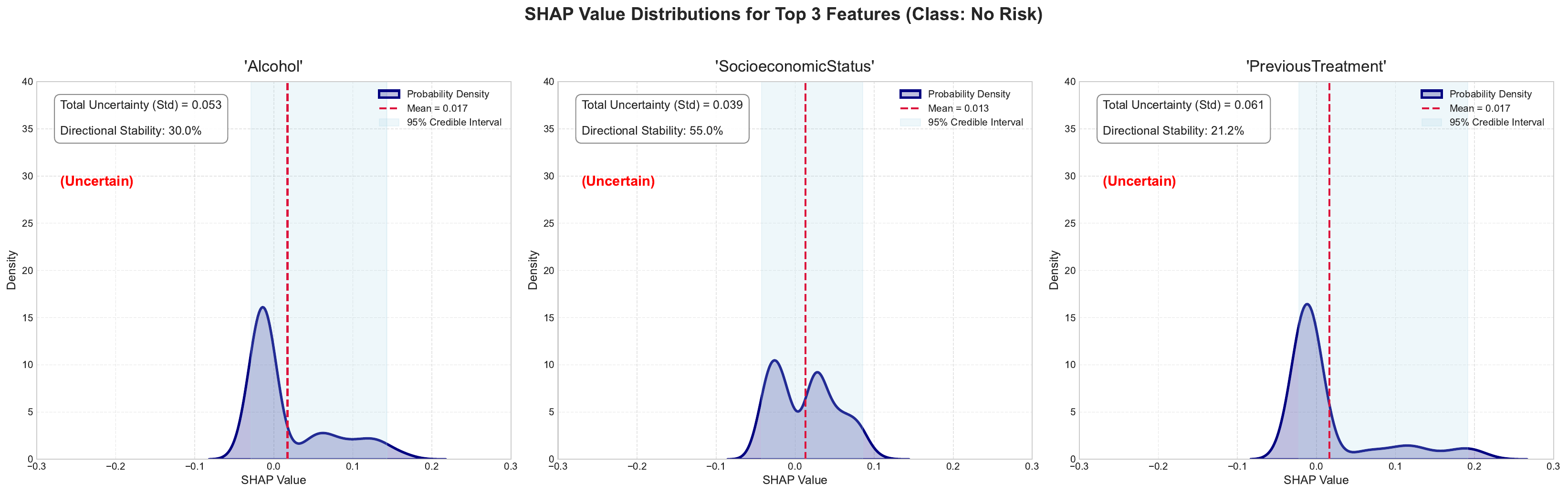}
    \caption{}
    \label{fig:ovarian_cancer_no_risk_shap_distribution}
\end{figure}
\begin{figure}[h!]
        \includegraphics[width=\linewidth]{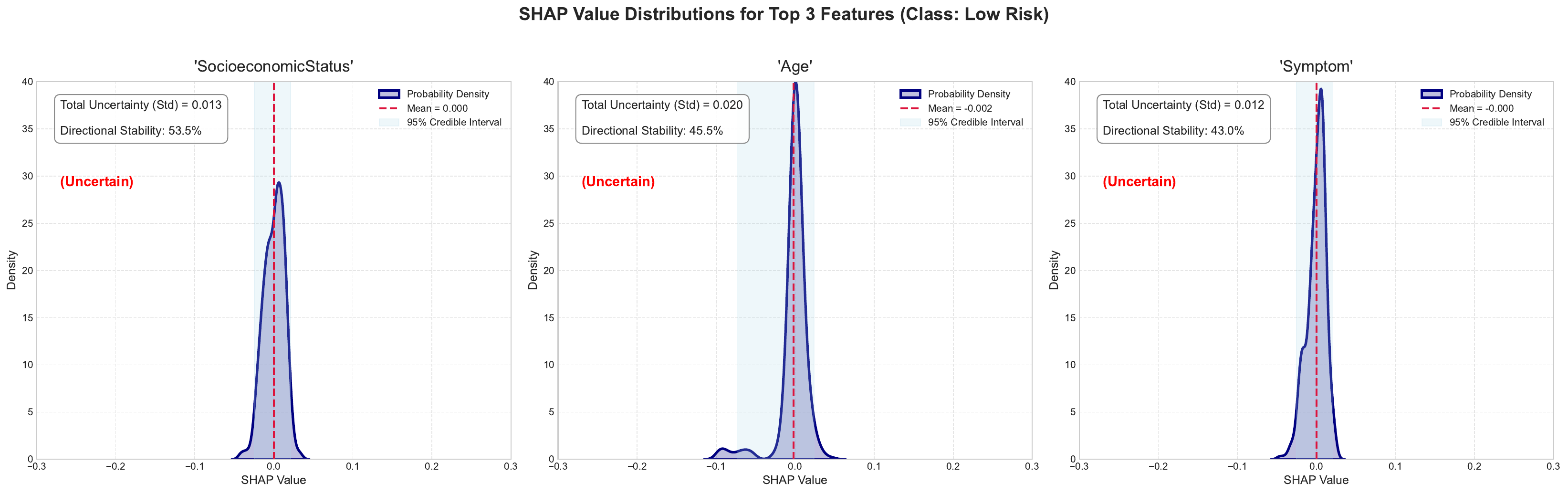}
    \caption{}
    \label{fig:ovarian_cancer_low_risk_shap_distribution}
\end{figure}
\begin{figure}[h!]
        \includegraphics[width=\linewidth]{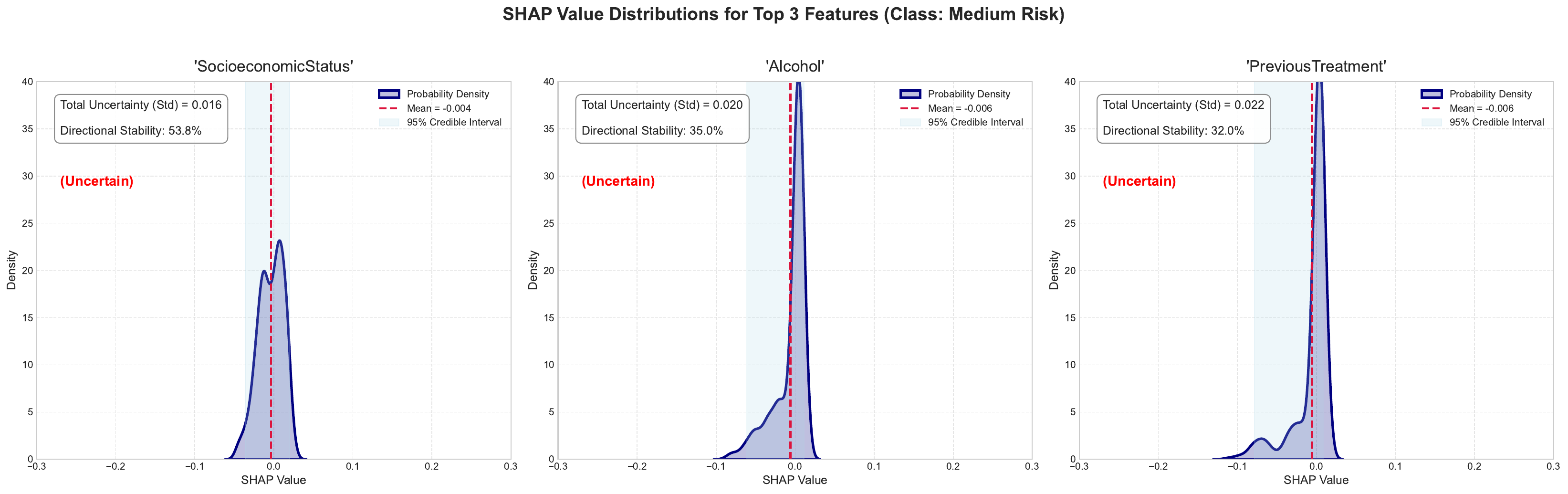}
    \caption{}
    \label{fig:ovarian_cancer_medium_risk_shap_distribution}
\end{figure}
\begin{figure}[h!]
        \includegraphics[width=\linewidth]{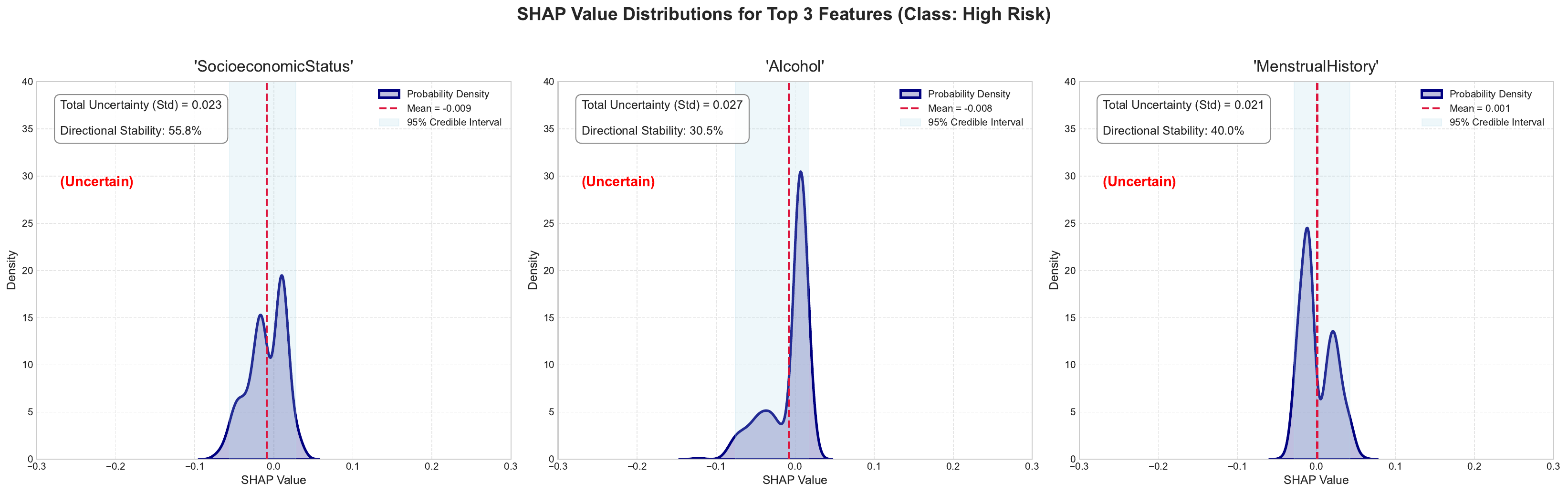}
    \caption{}
    \label{fig:ovarian_cancer_high_risk_shap_distribution}
\end{figure}
The Ovarian Cancer Dataset~\cite{ibe2025ovariancancer} contains 200,100 patient records collected hourly between January 2019 and December 2024. This highly detailed longitudinal dataset is useful for monitoring ovarian cancer risk and progression. It is designed to support prognostic modeling and progression risk assessment in ovarian cancer patients. This dataset contains 200,100 data points and 34 features. It was used to categorize ovarian cancer in females into risk categories. For each data point, the associated feature, \textit{Risk Label}, has four classes: \textit{No Risk, Low Risk, Medium Risk, and High Risk}. \textit{No Risk} coressponds to no evidence or indication of risk. \textit{Low Risk} corresponds to minimal probability of an adverse outcome or malignancy. \textit{Medium Risk} coressponds to moderate chance of risk; requires monitoring or further evaluation. \textit{High Risk} indicates a high probability of an adverse outcome or malignancy and likely warrants intervention. The description of the features we have discussed in the results are as mentioned further. \textit{Symptom} coressponds to clinical signs or patient-reported symptoms that are associated with the progression or risk of ovarian cancer and are used to characterize the disease. \textit{PreviousTreatment} coressponds to information on any medical therapies or interventions the patient underwent before the current assessment, such as chemotherapy, surgery, or radiation therapy. \textit{EnhancementPattern} is an imaging-derived feature that describes the contrast enhancement patterns observed in diagnostic scans. \textit{Smoking} is the patient's smoking history or status is a known risk factor impacting ovarian cancer progression and overall health. \textit{HormoneTherapy/HormoneTreatment} are the records of hormonal treatments received by the patient, including exogenous hormone administration, which may influence cancer risk or progression. \textit{SocioeconomicStatus} is a categorical or continuous measure reflecting the patient’s social and economic circumstances, which can affect access to healthcare and outcomes. \textit{SNP\_Status} is a genetic feature indicating the presence or absence of specific single nucleotide polymorphisms (SNPs) related to ovarian cancer susceptibility or progression.
The dataset was highly imbalanced, with the \textit{No Risk} class having 119,965 data points, the \textit{Low Risk} class having 40,092 data points, the \textit{Medium Risk} class having 30,068 data points, and the \textit{High Risk} class having 9,975 data points. 
\paragraph{For the reproducibility of these experiments, the following experimental details are provided: }The Ovarian Cancer dataset was preprocessed by standardizing column names and removing \textit{Timestamp} and \textit{ProgressionProbability} to prevent data leakage and to study the risk category instead of risk progression. The numerical features were imputed using the column median and scaled with \textit{StandardScaler}. Categorical features were encoded via \textit{LabelEncoder}. The preprocessed data was then subjected to a 70:30 stratified split based on the 4-class target variable, \textit{RiskLabel}. To address class imbalance, the training set was resampled using the \textit{SMOTETomek} algorithm with default parameters (\textit{k\_neighbors=5}) and \textit{random\_state=42}. A RF Classifier (\textit {n\_estimators=20, max\_depth=6, random\_state=42}) was trained on this resampled data. The performance of the model was evaluated using a test set that was not seen during the training process using F1 Score which was 55.7\%
The degree of uncertainty in SHAP explanations was studies using a separate RF Classifier (\textit{n\_estimators=20, class\_weight='balanced'}). The out-of-bag (OOB) accuracy of each of the 20 trees was utilized as performance-based weights. A Dirichlet distribution with a concentration parameter of $\alpha=$1.0 was employed to generate 20 posterior samples from the aforementioned weights. For each of the four target classes, SHAP values were computed across all 20 posterior samples. The resulting distributions were aggregated to calculate 95\% credible intervals and to decompose the total uncertainty of feature attributions into their aleatoric, epistemic, and entanglement components.

For the \textit{No Risk} class, the top three most important features were identified to be \textit{Alcohol, SocioeconomidStatus, \& PreviousTreatment}(Figure~\ref{fig:ovarian_cancer_shap}). For the \textit{Low Risk} class, the top three most important features were identified to be \textit{SocioeconomidStatus, Age, \& Symptom}. For the \textit{MediumRisk} class the top three most important features were identified to be \textit{SocioeconomidStatus, Alcohol, \& PreviousTreatment}. For the \textit{High Risk} class the top three most important features were identified to be \textit{SocioeconomicStatus, Alcohol, \& MenstrualHistory}. The results from the absolute mean SHAP values chart of the fifteen features having highest SHAP contributions for each class,(Figure~\ref{fig:ovarian_cancer_shap}) indicate that SHAP feature importance varies considerably among the underlying models. This suggests considerable variability across different model subsets. While the model consistently relies on features such as \textit{Alcohol, Symptom, MenstrualHistory, SocioeconomicStatus, etc.} , it does so with substantial epistemic ambiguity regarding their precise impact magnitude. In contrast, features such as \textit{MenopauseAge} (No Risk Class) exhibit low average SHAP values and narrow uncertainty narrow violin plots, reflecting low importance and high confidence in their negligible contribution. This hints at stable but marginal roles in predicting the different classes. One major difference observed is in terms of positively or negatively influencing predictions.

We also studied the top three features of this dataset and their associated epistemic uncertainty, sign stability, and SHAP value distribution, as well as the mean SHAP and 95\% SHAP confidence interval. For the \textit{No Risk} category, the epistemic uncertainties for the features \textit{No Risk} category the epistemic uncertainties for the features \textit{Alcohol, SocioeconomicStatus, PreviousTreatment} are 0.053, 0.039, and 0.0561, respectively (Figure~\ref{fig:ovarian_cancer_no_risk_shap_distribution}). The sign stabilities for these features are 30.0\%, 55.0\%, and 21.2\%, respectively. These metrics reflect model variance in attribution for each feature. The low standard deviation of epistemic uncertainties in the top three features suggests consistent SHAP values across the ensemble and reflects certainty about the features' influence. The SHAP distribution is non-uniform and skewed, signaling high information uncertainty about the feature's impact. These features have low sign stability, indicating interpretive inconsistency and low epistemic uncertainty. For the \textit{Low Risk} category, the epistemic uncertainties for the features \textit{SocioeconomicStatus, Age, Symptom} are 0.013, 0.026, and 0.012, respectively. The sign stabilities for these features are 53.5\%, 45.5\%, and 43.0\%, respectively (Figure:~\ref{fig:ovarian_cancer_low_risk_shap_distribution}). These metrics reflect model variance in attribution for each feature. The low standard deviation of epistemic uncertainties in the top three features suggests consistent SHAP values across the ensemble and reflects certainty about the features' influence. The SHAP has a uniform distribution which is converging to the mean SHAP value for each individual features for the \textit{SocioeconomicStatus, Age, \& Symptom}  features, signaling high information uncertainty about the feature's impact. The features indicate low interpretive consistency despite low epistemic uncertainty. The SHAP distribution exhibits flatness, indicating disagreement across the sub-ensembles. For the \textit{Medium Risk} category, the epistemic uncertainties for the features \textit{SocioeconomicStatus, Alcohol, \& PreviousTreatment} are 0.016, 0.020, and 0.022, respectively. The sign stabilities for these features are 53.8\%, 35.0\%, and 32.0\%, respectively (see Figure:~\ref{fig:ovarian_cancer_medium_risk_shap_distribution}). These metrics reflect model variance in attribution for each feature. The low standard deviation of epistemic uncertainties in the top three features suggests consistent SHAP values across the ensemble and reflects certainty about the features' influence. The SHAP has a non-uniform distribution for the features, signaling high information uncertainty about the feature's impact as evident in previous classes as well. The SHAP distribution exhibits converging peaks at the means SHAP values. For the \textit{High Risk} category, the epistemic uncertainties for the features \textit{SocioeconomicStatus, Alcohol, MenstrualHistory} are 0.023, 0.027, and 0.021, respectively. The sign stabilities for these features are 55.8\%, 30.5\%, and 40.0\%, respectively (see Figure:~\ref{fig:ovarian_cancer_high_risk_shap_distribution}). These metrics reflect model variance in attribution for each feature. The low standard deviation of epistemic uncertainties in the top three features suggests consistent SHAP values across the ensemble and reflects certainty about the features' influence. The SHAP distribution is skewed with multiple peaks for all the features, signaling high information uncertainty about the feature's impact.

\begin{figure}[h!]
    \centering
    \includegraphics[width=\textwidth]{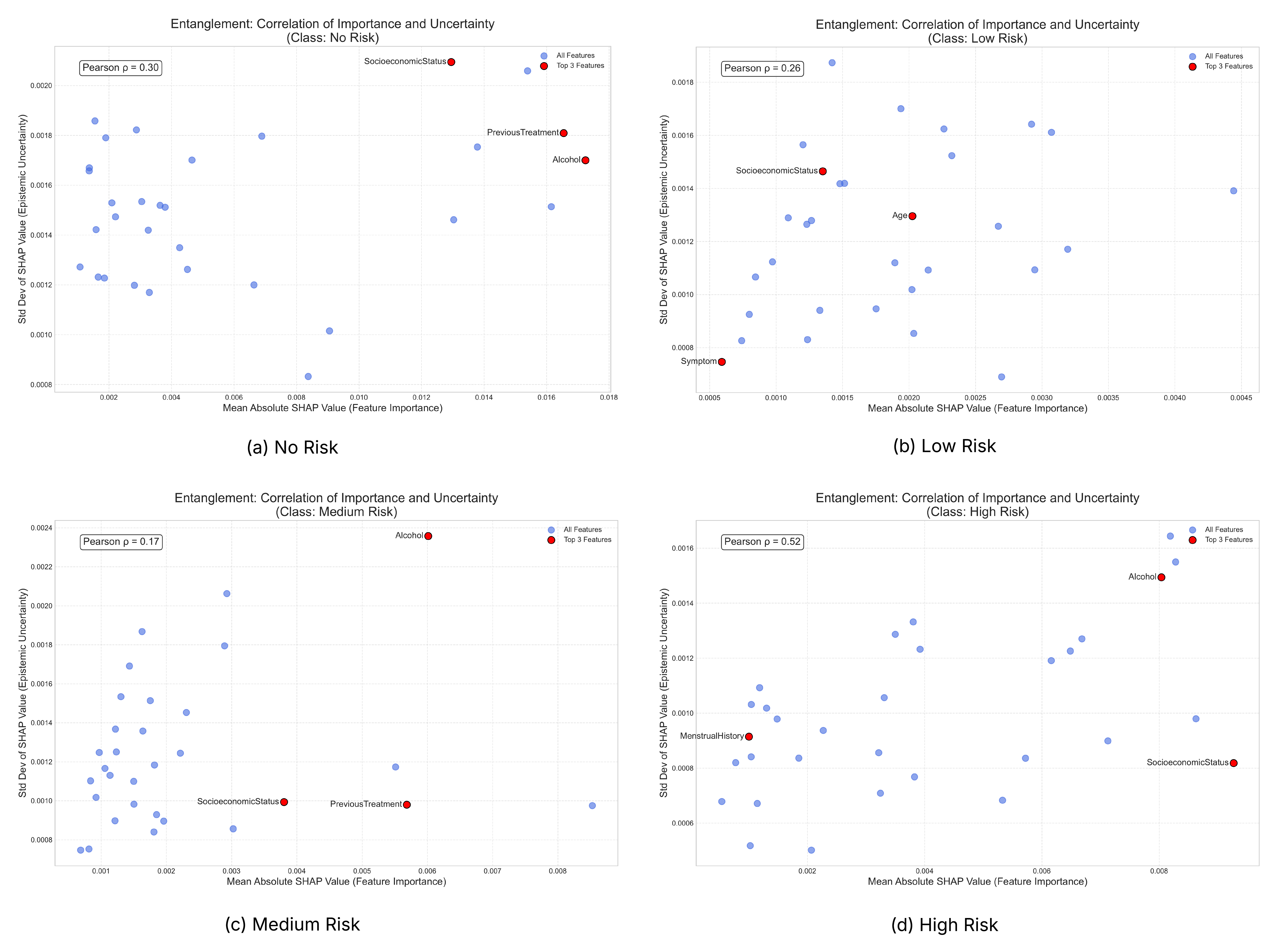}
    \caption{}
    \label{fig:ovarian_entanglement}
\end{figure}
The entanglement analysis (See Figure:~\ref{fig:ovarian_entanglement}) demonstrates a significant and robust correlation with the predicted risk category. The correlation is most robust for the high-risk class, Pearson correlation ($\rho$) is 0.52, suggesting that the model's most salient features for high-risk predictions (i.e., alcohol consumption, menstrual history, and socioeconomic status) are also its most unstable. This correlation progressively weakens for the \textit{No Risk} ($\rho=$0.30), \textit{Low Risk}  ($\rho=$0.26), and \textit{Medium Risk} ($\rho=$0.17) classes. This pattern indicates that model uncertainty is disproportionately concentrated in the high-stakes predictions where reliability is most critical. The entanglement analysis reveals a critical technical flaw in the model's explainability: a strong, positive correlation between feature importance and epistemic uncertainty that is most pronounced in the \textit{High Risk} prediction class ($\rho=$0.52). This indicates that the model's most significant features, which are crucial for making high-stakes predictions, demonstrate the highest variability in their SHAP values across different model instances. This finding suggests that the underlying reasoning for these predictions is inherently unstable. Conversely, lower-risk classes exhibit diminished correlations, with the medium-risk class ($\rho=$0.17) exhibiting a decoupled pattern where uncertainty is diffuse rather than associated with specific features. This class-dependent entanglement indicates that the model's reliability is not uniform, but is severely compromised precisely where robust and stable explanations are most critical for trustworthy deployment. In conclusion, reliability of explanations is class-dependent, with the least stable explanations being generated for the most critical high risk patients. This results in a substantial increase in the operational risk, as the rationale for categorizing a patient as high-risk becomes questionable. The model's reasoning is more stable and consistent for lower-risk predictions. In regard to its deployment, this necessitates a degree of caution when employing the model's explanations for high-risk decisions. It also underscores the necessity for refining the model with a specific focus on stabilizing the features that are instrumental in producing its most critical predictions. 
This analysis reveals a significant weakness in the model's application for ovarian cancer risk assessment: its explanatory reliability is critically dependent on the predicted risk category. The visualizations, particularly the entanglement analysis, were instrumental in reaching this conclusion. They demonstrate a strong positive correlation between a feature's importance and its uncertainty. This problem is most pronounced in the high-risk class ($\rho$ = 0.52). This means that the features on which the model relies most for its most critical predictions, such as \textit{Alcohol, Socioeconomic Status, \& Menstrual History}, are also the least stable and reliable in their explanatory power.
This finding is important for clinical decision-making because it calls into question the trustworthiness of the model's reasoning at the most critical instances. While the model may correctly predict that a patient is high risk, the instability of the underlying feature attributions means clinicians cannot be confident in the reasoning behind that conclusion. This introduces substantial operational risk because the rationale for a high-stakes diagnosis is questionable. Therefore, for high-risk predictions, the model's explanations cannot be taken at face value and require careful human oversight. To address the feedback, the domain-expert-in-the-loop method could be used to pre-process the data, extensive data collection, etc.
\subsection*{SEER Breast Cancer Dataset}
\begin{figure}[h!]
    \centering
    \includegraphics[width=\textwidth]{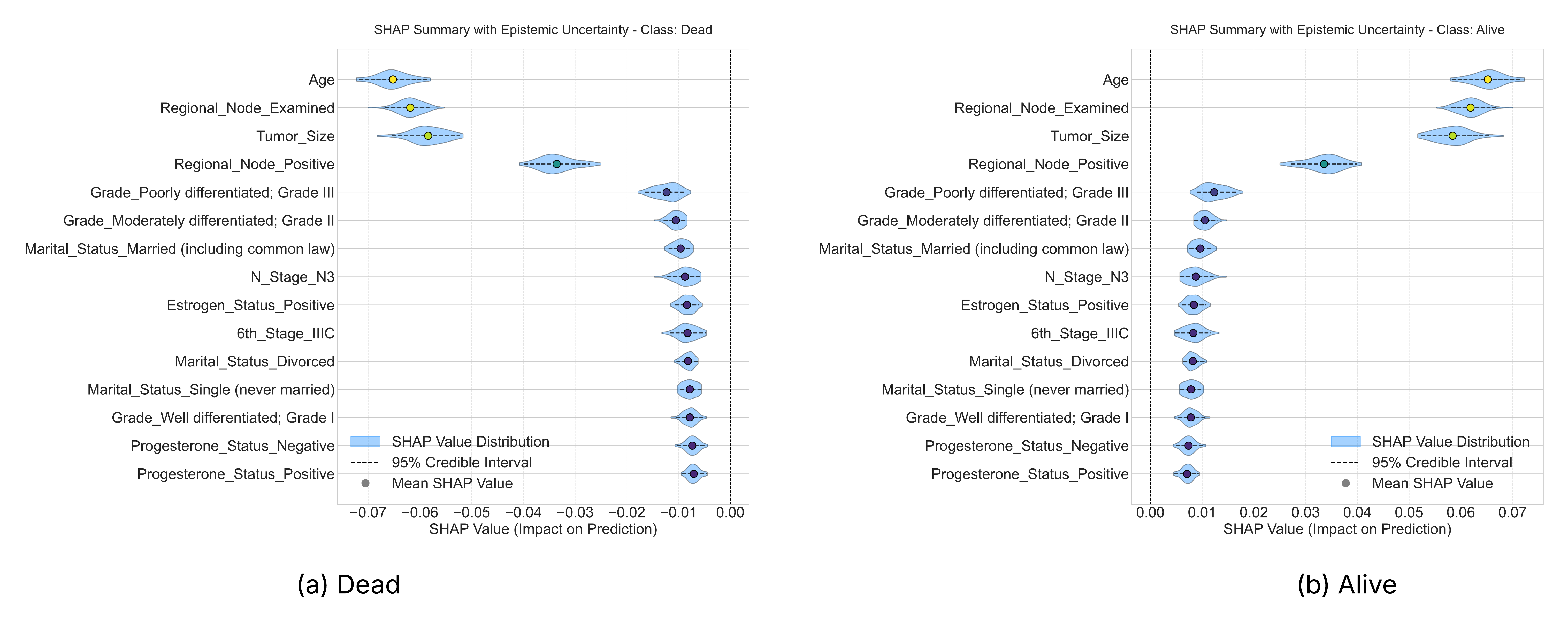}
    \caption{}
    \label{fig:seer_alive_breast_cancer_shap}
\end{figure}
\begin{figure}[h!]
    \centering
        \includegraphics[width=\linewidth]{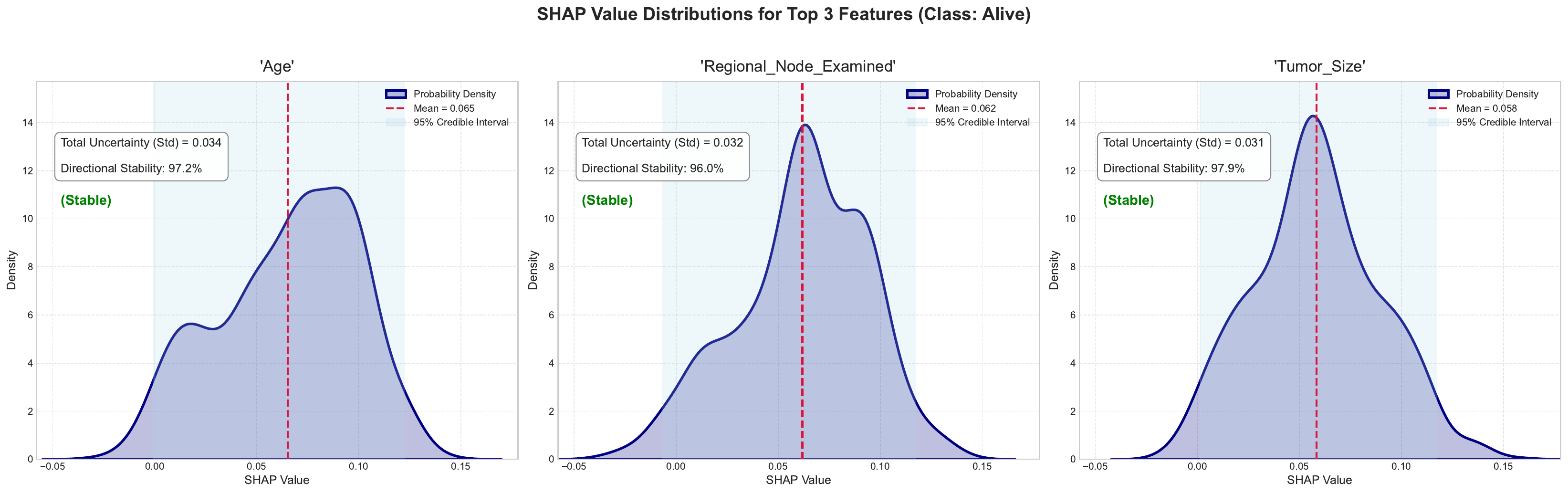}
    \caption{}
\label{fig:seer_alive_breast_cancer_shap_distribution}
\end{figure}
\begin{figure}[h!]
    \centering
        \includegraphics[width=\linewidth]{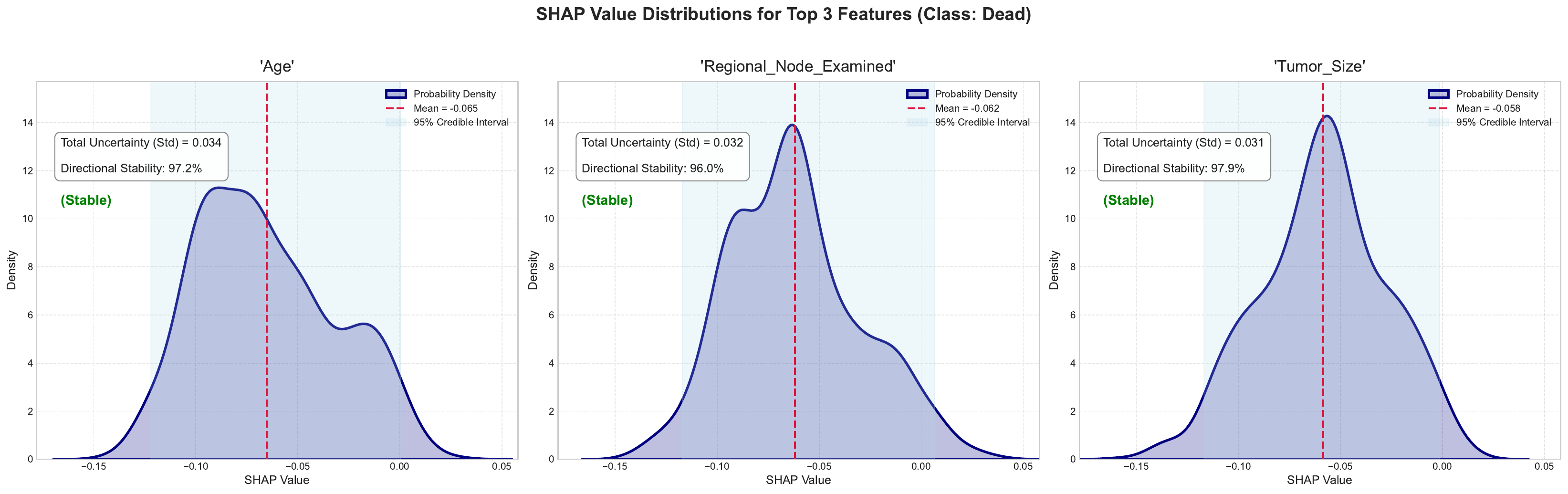}
    \caption{}
\label{fig:seer_dead_breast_cancer_shap_distribution}
\end{figure}
The SEER Breast Cancer Dataset~\cite{seerbreast} is a comprehensive cancer registry that provides extensive information on breast cancer cases, including patient demographics, tumor characteristics, treatment details, and survival outcomes. The utilization of this method is prevalent in the analysis of treatment outcomes, as it captures comprehensive, population-based longitudinal data. This capability enables researchers and clinicians to assess the efficacy of diverse interventions across a range of patient populations and clinical settings. For instance, advanced predictive models developed using SEER data apply machine learning to predict individual patient survival and treatment response, enabling personalized treatment decisions and optimizing treatment strategies to extend survival while minimizing adverse effects. This capability is of particular significance in complex cases, such as metastatic breast cancer, where the guidance provided by clinical trials is limited and SEER-based models facilitate decision-making by simulating the outcomes of various treatment options that are tailored to the unique characteristics of each patient and tumor~\cite{manikandan2023integrative, ren2024developing}. The dataset contains 4,024 data points, 16 features, and associated survival outcomes, specifically the \textit{Alive and Dead} status, for female patients diagnosed with breast cancer. 

The \textit{Alive} class indicates the patients who were alive at the final follow-up or censoring time. \textit{Dead} class indicates patients who died from breast cancer or any other cause during the observation period. The features under discussion are described further text. 
\textit{Age}: The age of the patient at the time of the initial breast cancer diagnosis. This influences prognosis and treatment choice.
\textit{Regional\_Node\_Positive}: The number of regional lymph nodes confirmed positive for cancer, indicating spread.
\textit{Grade\_Poorly\ differentiated\ (Grade\ III)}: Tumor cells are highly abnormal and grow/spread aggressively.
\textit{Grade\_Moderately\ differentiated\ (Grade\ II)}: Tumor cells are highly abnormal and grow/spread aggressively.
\textit{Regional\_Node\_Examined} indicating that the total number of regional lymph nodes examined for cancer involvement.
\textit{Progesterone\_Status\_Negative} indicating that the tumor cells lack progesterone receptors, which may affect response to hormone therapy.
\textit{T\_Stage\_T4} corresponding that the tumor has invaded the chest wall and/or skin (advanced tumor size/invasion stage).
\textit{N\_Stage\_N1} denoting that the cancer has spread to one to three axillary lymph nodes.
\textit{N\_Stage\_N2} indicating that the cancer has spread to four to nine axillary lymph nodes, or the nodes are fixed or matted.\textit{N\_Stage\_N3} indicating that the cancer has spread to 10 or more axillary nodes, as well as to the infraclavicular or internal mammary nodes.
\textit{6th\_Stage\_IIA} is the AJCC 6th edition Stage IIA indicating moderately advanced local disease.\textit{6th\_Stage\_IIB} is the AJCC 6th edition Stage IIB indicating larger tumor size and/or more node involvement.\textit{6th\_Stage\_IIIA} is the AJCC 6th edition Stage IIIA indicating advanced local spread to several regional nodes.\textit{6th\_Stage\_IIIB} is the AJCC 6th edition Stage IIIB indicating tumor involves chest wall or skin and may involve nodes.\textit{6th\_Stage\_IIIC} is the AJCC 6th edition Stage IIIC indicating extensive lymph node involvement near collarbone or breastbone.

\paragraph{For the reproducibility of these experiments, the following experimental details are provided: }
The target variable, \textit{Status}, which indicates patient survival, was encoded as integers ("Alive" or "Dead") using \textit{LabelEncoder}. The fully preprocessed dataset was then split into a training set (70\%) and a testing set (30\%). A stratified split was employed to ensure consistent proportions of survival outcomes in both the training and testing data. A RF Classifier was trained on the processed data. The model was configured with 50 estimators (\textit{n\_estimators=50}). To address the class imbalance of the target variable, the \textit{class\_weight="balanced"} parameter was used. The \\ \textit{oob\_score=True} parameter was also enabled to estimate the model's performance using out-of-bag samples. For consistency, the random state was fixed at 42. After training, the model achieved an 84.4\% micro-averaged F1 score on the unseen test set. To evaluate epistemic uncertainty in the model's SHAP-based explanations, a Dirichlet process was implemented. The out-of-bag (OOB) accuracy of each of the 50 trees in the random forest framework was calculated to serve as a performance-based weight. A Dirichlet distribution was used to generate 50 posterior samples (n\_dirichlet\_samples = 50). The concentration parameter, $\alpha$, was set to 0.7. When $\alpha$ < 1, the sampling process is biased to give a higher probability to trees with better OOB performance. This focuses the uncertainty analysis on the most competent parts of the ensemble. For each target class, the full suite of uncertainty analyses was performed. This involved computing SHAP values for each of the 50 posterior samples and aggregating the results to produce distributions, credible intervals, and decompositions of uncertainty into aleatoric, epistemic, and entanglement components for the most impactful features.  The implementation of the proposed framework on the test data set was undertaken to assess the epistemic uncertainty inherent in predictions pertaining to test data.
The absolute mean SHAP values chart (Figure:~\ref{fig:seer_alive_breast_cancer_shap}) shows the fifteen features with highest feature importance and their SHAP feature importance varies considerably among the top contributing features in the underlying models. This suggests considerable variability across different model sub-ensembles. While the model consistently relies on features such as \textit{Age, Regional\_Node\_Examined, Tumor\_Size, Regional\_Node\_Positive}, the model exhibits substantial epistemic ambiguity regarding their precise impact magnitude. In contrast, features such as \textit{Grade\_Poorly differentiated: Grade III, Grade\_Poorly differentiated: Grade II, Progesterone\_Status\_Negative, Progesterone\_Status\_Positive, etc.,} exhibit low average SHAP values and narrow uncertainty violin plots. This reflects both low importance and high confidence in their negligible contribution, hinting at stable but marginal roles in predicting the different classes. One major difference observed is in terms of positively or negatively influencing predictions. For each survival status, we calculate the absolute SHAP values. For the  \textit{Dead or Deceased} class, the top three most important features were identified as \textit{Age, Regional\_Node\_Examined, \&} \textit{Tumor\_Size}. For the \textit{Surviving or Alive} class, the top three most important features are \textit{Age, Regional\_Node\_Examined, \&} \textit{Tumor\_Size} (Figure:~\ref{fig:seer_alive_breast_cancer_shap}). We also examined the top three features of this dataset and their associated epistemic uncertainty, sign stability, and SHAP value distribution with mean SHAP and 95\% SHAP confidence intervals. For the \textit{Surviving or Alive} category, the epistemic uncertainties for the features \textit{Age, Regional\_Node\_Examined, \&} \textit{Tumor\_Size} are 0.034, 0.032, and 0.031, respectively. The sign stabilities for these features are 97.2\%, 96.0\%, and 97.9\%, respectively (Figure:~\ref{fig:seer_alive_breast_cancer_shap_distribution}). For the \textit{Deceased or Dead} category, the epistemic uncertainties for the features \textit{Age, Regional\_Node\_Examined,} \& \\ \textit{Tumor\_Size}  are 0.034, 0.032, and 0.031, respectively (Figure ~\ref{fig:seer_dead_breast_cancer_shap_distribution}). The sign stabilities for these features are 97.2\%, 96.0\%, and 97.9\%, respectively. For both classes, the metric reflects model variance in attribution for each feature. The low standard deviation of epistemic uncertainties in the top three features suggests consistent SHAP values across the ensemble and reflects certainty about the features' influence. SHAP values are non-uniform for \textit{Age, Regional\_Node\_Examined, \&} \textit{Tumor\_Size} features, signaling high uncertainty about the features' impact, along with a relatively dispersed distribution. This reinforces the model's uncertainty about precise attribution. The sign stabilities indicate high consistency with low epistemic uncertainty. 

\begin{figure}[h!]
    \centering
    \includegraphics[width=\textwidth]{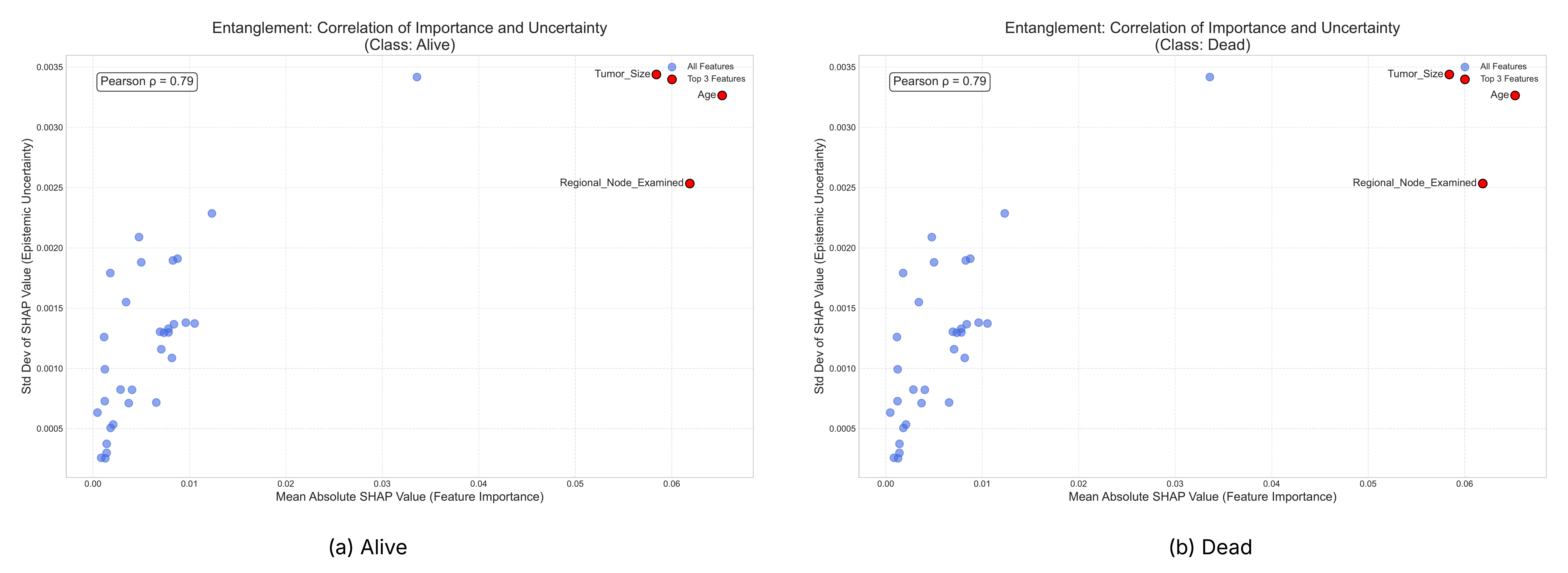}
    \caption{}
    \label{fig:seer_entanglement}
\end{figure}
The entanglement analysis (See Figure:~\ref{fig:seer_entanglement}) reveals a robust and consistent positive Pearson correlation ($\rho$) 0.79 between a feature's importance and its explanation uncertainty for both the \textit{Alive} and \textit{Dead} patient outcomes. This finding suggests that the most critical clinical factors driving the model's predictions, specifically \textit{Tumor\_Size}, \textit{Age}, and \textit{Regional\_Node\_Examined} are also the factors with the most unstable explanations. It is imperative to acknowledge that for features not associated with importance, the epistemic uncertainty (standard deviation of SHAP) was minimal and directional stability was substantial. This indicates that the model consistently allocated a negligible and consistent effect to these features. Consequently, the high uncertainty is concentrated exclusively on the most significant drivers of the prediction. The conclusion drawn from this analysis is that this entanglement is a fundamental property of the model, with implications for predictions of both survival and mortality. The instability is not random but systematically targets the core of the model's reasoning. This results in a substantial trust barrier, as the rationale for a high-risk prediction is not a constant, dependable fact but rather exhibits significant variability. In the context of deployment, the model's rationale is deemed unreliable for its most critical predictions, necessitating the collection of additional data or the refinement of the model to stabilize its processing of these pivotal features.

\section*{DISCUSSION}
This research study decomposes SHAP into two categories of uncertainty quantification: aleatoric and epistemic. Breaking down uncertainty allowed us to determine whether it arises from uneliminatable noise in the data or from a lack of knowledge about the true data/model distribution. We also introduced an entanglement term that captures the interaction or covariance between data and model uncertainties. 

Our approach provides a deeper understanding of SHAP uncertainty in terms of intervals and enables investigation of its origin. In domains such as healthcare, where the consequences can be significant, our method is useful. The proposed approach captures uncertainties that simple intervals cannot while aligning with modern uncertainty quantification practices. DST quantifies both certainty ($Bel$) and possibility ($Pl$) for SHAP attributions, allowing for the explicit modeling of ignorance and epistemic uncertainty. This is particularly useful in ensemble bagging models with conflicting feature attributions. DST offers a non-probabilistic method of expressing confidence in SHAP values, which is useful when data is scarce or evidence is subjective. The framework extends beyond classical probability, offering tools for interpreting and managing uncertainty in model explanations. Identifying constrained SHAP intervals is often challenging in practice for large or complex models~\cite{napolitano2024evaluating}. The proposed framework aims to facilitate the estimation of simple uncertainty intervals by leveraging the structural properties of belief functions, uncertainty theorems, and Dirichlet processes. These processes can be efficiently sampled or approximated, thereby enhancing the framework's efficacy and accessibility. The framework provides control parameters, such as $\alpha$, which enable users to control the exploration-exploitation process of the hypothesis space. Users can choose balanced exploration, exploration of the best-performing trees, or uniform uncertainty estimation. Furthermore, uncertainty theory quantifies confidence in attribution magnitude, with entropy minimization guiding optimal data acquisition. This could facilitate explanations of SHAP values that account for uncertainty, even in high-dimensional or real-world settings. For high-stakes domains such as healthcare, the UbiQTree could be combined with recent model-agnostic approaches such as Counterfactual Paths (CPATH)~\cite{pfeifer2025explaining} as well, a model-agnostic approach to providing global explanations for machine learning (ML) models, particularly those trained on tabular data. CPATH overcomes the limitations of traditional feature importance methods using a graphical, path-based approach inspired by counterfactual explanations. CPATH identifies sequential feature permutations that significantly influence changes in model predictions and incorporates domain knowledge graphs to restrict the feature space. This results in more targeted and interpretable explanations. Integrating UbiQTree with CPATH could provide a richer explainability framework in healthcare by combining uncertainty quantification with global, structural counterfactual reasoning. UbiQTree decomposes uncertainties in tree ensemble predictions for a patient's data, indicating which features and regions of the feature space the model is confident about. CPATH constructs counterfactual paths via sequential feature permutations guided by a knowledge graph to understand which combinations of features more globally drive class changes in the model's behavior.

\section*{CONCLUSION}
Our framework enhances uncertainty reporting. In a healthcare context, for example, SHAP values with 95\% confidence intervals replace point estimates. The confidence-triggered verification is essential for determining which features require a domain expert's review. This work also touches on the boundaries of AI regulations which emphasizes on the fact that the AI frameworks must provide auditable uncertainty metrics~\cite{dubey2024nested}. Together, variance, entropy, and sign stability provide a complete picture of uncertainty. Each chart provides insight into the stability of a SHAP attribution. The user can examine standard deviation, entropy, and sign stability. A mean $\pm2\sigma$ SHAP indicates high epistemic uncertainty if SHAP varies widely across sub-models. The SHAP confidence interval represents the distribution of SHAP values; multimodality or flatness indicates disagreement across model variants. Uncertainty metrics quantify the reliability and stability of a feature's attribution. 




\newpage


\section*{RESOURCE AVAILABILITY}


\subsection*{Lead contact}


Requests for further information and resources should be directed to and will be fulfilled by the lead contact, Akshat Dubey (DubeyA@rki.de).

\subsection*{Materials availability}

Given the nature of the research, no materials (chemical or physical) are produced or generated.

\subsection*{Data and code availability}


\begin{itemize}
    \item All original code has been deposited at Zenodo under the DOI 10.5281/zenodo.17542114 and is publicly available as of the date of publication.\cite{akshat_dubey_2025_17542114}
    \item Any additional information required to reanalyze the data reported in this paper is available from the lead contact upon request.    
\end{itemize}

\section*{ACKNOWLEDGMENTS}
One of the authors of this work has been financially supported by the German Federal Ministry of Health (BMG) under grant No.: ZMI5- 2523GHP027 (project “Strengthening National Immunization Technical Advisory Groups and their Evidence-based Decision-making in the WHO European Region and Globally” SENSE) part of the Global Health Protection Programme, GHPP. We also thank the Center for Artificial Intelligence in Public Health Research (ZKI-PH) at the Robert Koch Institute and the Department of Mathematics \& Computer Science at the Free University of Berlin for the additional support.

\section*{AUTHOR CONTRIBUTIONS}

Conceptualization, A.D.; methodology, A.D. and A.A. and G.H.; experimentations, A.D. and A.A. and B.İ.; analysis, A.D. and A.A. and B.İ.; writing-–original draft, A.D. and A.A. and B.İ. and G.H.; writing-–review \& editing, A.D. and A.A. and B.İ. and G.H.; funding acquisition, B.İ and G.H.; supervision, A.A. and B.İ. and G.H., project administration, G.H.

\section*{DECLARATION OF INTERESTS}

The authors declare no competing interests.

\newpage
\section*{FIGURE TITLES}

\begin{itemize}
    \item Figure 1: The SHAP summary plot provides a visual representation of the impact of various features (fifteen most contributing features) on a model's prediction for the \textit{Short Stay, Medium Stay, \& Long Stay} class, incorporating epistemic uncertainty. The violin plots illustrate the distribution of SHAP values for each feature, with individual hypothesis samples represented by gray points. The blue shaded area indicates the 95\% confidence interval($\pm2\sigma$) along with the horizontal dashed line of the feature's impact on the prediction. The plot suggests that \textit{NumTransfers, NumNotes, NumCallouts} are the most impactful feature, significantly contributing to the high probability of the prediction's shift towards the \textit{Short Stay} class; \textit{NumTransfers, NumCallouts, NumNotes} are the most impactful feature, significantly contributing to the high probability of the prediction's shift toward the \textit{Medium Stay} class; \textit{NumTransfers, NumNotes, NumCallouts} are the most impactful feature, significantly contributing to the high probability of the prediction's shift toward the \textit{Long Stay} class The features with mean SHAP value on the right side of the vertical dashed line contribute positively towards the prediction and vice-versa.
    \item Figure 2: The distribution of SHAP values for the four most contributing features to the \textit{Short Stay} class was examined to further investigate the stability and epistemic uncertainty of the features. The KDE plot shows the distribution of SHAP values collected from different model samples. The red dashed vertical line marks the mean SHAP value, and the shaded region represents the 95\% credible interval.

    \item Figure 3: The most contributing features to the \textit{Medium Stay} class was examined to further investigate the stability and epistemic uncertainty of the features. The KDE plot shows the distribution of SHAP values collected from different model samples. The red dashed vertical line marks the mean SHAP value, and the shaded region represents the 95\% credible interval.
    
    \item Figure 4: The distribution of SHAP values for the four most contributing features to the \textit{Long Stay} class was examined to further investigate the stability and epistemic uncertainty of the features. The KDE plot shows the distribution of SHAP values collected from different model samples. The red dashed vertical line marks the mean SHAP value, and the shaded region represents the 95\% credible interval.
    
    \item Figure 5: This figure illustrates the relationship between a feature's importance, measured by the mean absolute SHAP value, and the uncertainty surrounding its explanation, quantified by the standard deviation of SHAP values across a model ensemble, for various model prediction classes for the MIMIC-III Dataset. A high positive correlation ($\rho$) indicates that the model's most significant features are also the most unstable, meaning their attributed contribution to the prediction varies significantly. Conversely, a low correlation suggests that feature importance and explanation stability are decoupled; the model can have important features it is certain about, and unimportant features it is uncertain about. This analysis is important for assessing the reliability of a model's explanations, especially in high-stakes contexts.
    
    \item Figure 6: The SHAP summary plot provides a visual representation of the impact of the fifteen features with highest SHAP feature importance on a model's prediction for the \textit{No Risk, Low Risk, Medium Risk, \& High Risk} class, incorporating epistemic uncertainty. The violin plots illustrate the distribution of SHAP values for each feature, with individual hypothesis samples represented. The blue shaded area indicates the 95\% confidence interval($\pm\sigma$) of the feature's impact on the prediction. The plot suggests that \textit{Alcohol, SocioeconomicStatus, PreviousTreatment} are the most impactful feature, significantly contributing to the high probability of the prediction's shift toward the \textit{No Risk} class; \textit{SocioeconomicStatus, Age, Symptom} are the most impactful feature, significantly contributing to the high probability of the prediction's shift toward the \textit{Low Risk} class; \textit{SocioeconomicStatus, Alcohol, PreviousTreatment} are the most impactful feature, significantly contributing to the high probability of the prediction's shift toward the \textit{Medium Risk} class; \textit{SocioeconomicStatus, Alcohol, MenstrualHistory} are the most impactful feature, significantly contributing to the high probability of the prediction's shift toward the \textit{High Risk} class.
    
    \item Figure 7: The distribution of SHAP values for the four most contributing features to the \textit{No Risk} class was examined to further investigate the stability and epistemic uncertainty of the features. The KDE plot shows the distribution of SHAP values collected from different model samples. The red dashed vertical line marks the mean SHAP value, and the shaded region represents the 95\% credible interval.
    
    \item Figure 8: The distribution of SHAP values for the four most contributing features to the \textit{Low Risk} class was examined to further investigate the stability and epistemic uncertainty of the features. The KDE plot shows the distribution of SHAP values collected from different model samples. The red dashed vertical line marks the mean SHAP value, and the shaded region represents the 95\% credible interval.
    
    \item Figure 9: The distribution of SHAP values for the four most contributing features to the \textit{Medium Risk} class was examined to further investigate the stability and epistemic uncertainty of the features. The KDE plot shows the distribution of SHAP values collected from different model samples. The red dashed vertical line marks the mean SHAP value, and the shaded region represents the 95\% credible interval.
    
    \item Figure 10: The distribution of SHAP values for the four most contributing features to the \textit{High Risk} class was examined to further investigate the stability and epistemic uncertainty of the features. The KDE plot shows the distribution of SHAP values collected from different model samples. The red dashed vertical line marks the mean SHAP value, and the shaded region represents the 95\% credible interval.
    
    \item Figure 11: This figure illustrates the relationship between a feature's importance, measured by the mean absolute SHAP value, and the uncertainty surrounding its explanation, quantified by the standard deviation of SHAP values across a model ensemble, for various model prediction classes for the Ovarian Cancer Dataset. A high positive correlation ($\rho$) indicates that the model's most significant features are also the most unstable, meaning their attributed contribution to the prediction varies significantly. Conversely, a low correlation suggests that feature importance and explanation stability are decoupled; the model can have important features it is certain about, and unimportant features it is uncertain about. This analysis is important for assessing the reliability of a model's explanations, especially in high-stakes contexts.)
    
    \item Figure 12: The SHAP summary plot provides a visual representation of the impact of various features (fifteen most contributing) on a model's prediction for the \textit{Deceased \& Alive} class, incorporating epistemic uncertainty. The violin plots illustrate the distribution of SHAP values for each feature, with individual hypothesis samples represented. The blue shaded area indicates the 95\% confidence interval($\pm\sigma$) of the feature's impact on the prediction along with the horizontal dashed line of the feature’s impact on the prediction. The plot suggests that \textit{Age, Regional\_Node\_Examined, Tumor\_Size} are the most impactful features, significantly contributing to the high probability of the prediction's shift towards both \textit{Deceased \& Alive} class, although in different directions. SHAP value on the right side of the vertical dashed line contribute positively towards the prediction and vice-versa.
    
    \item Figure 13: The distribution of SHAP values for the four most contributing features to the \textit{Alive} class was examined to further investigate the stability and epistemic uncertainty of the features. The KDE plot shows the distribution of SHAP values collected from different model samples. The dashed vertical line marks the mean SHAP value, and the shaded region represents the 95\% credible interval.
    
    \item Figure 14: The distribution of SHAP values for the four most contributing features to the \textit{Dead} class was examined to further investigate the stability and epistemic uncertainty of the features. The KDE plot shows the distribution of SHAP values collected from different model samples. The red dashed vertical line marks the mean SHAP value, and the shaded region represents the 95\% credible interval.
    
    \item Figure 15: This figure illustrates the relationship between a feature's importance, measured by the mean absolute SHAP value, and the uncertainty surrounding its explanation, quantified by the standard deviation of SHAP values across a model ensemble, for various model prediction classes for the SEER Breast Cancer Dataset. A high positive correlation ($\rho$) indicates that the model's most significant features are also the most unstable, meaning their attributed contribution to the prediction varies significantly. Conversely, a low correlation suggests that feature importance and explanation stability are decoupled; the model can have important features it is certain about, and unimportant features it is uncertain about. This analysis is important for assessing the reliability of a model's explanations, especially in high-stakes contexts.

\end{itemize}

\newpage

\bibliography{references}

\bigskip


\newpage

\end{document}